\documentclass{article}

\usepackage{arxiv}

\usepackage[utf8]{inputenc} % allow utf-8 input
\usepackage[T1]{fontenc}    % use 8-bit T1 fonts
\usepackage{hyperref}       % hyperlinks
\usepackage{url}            % simple URL typesetting
\usepackage{booktabs}       % professional-quality tables
\usepackage{amsfonts}       % blackboard math symbols
\usepackage{nicefrac}       % compact symbols for 1/2, etc.
\usepackage{microtype}      % microtypography
\usepackage{lipsum}
\usepackage{graphicx}
\graphicspath{ {./images/} }

\usepackage{amsmath} 
\usepackage{enumitem} % For customizing lists
\usepackage{float}

\usepackage{subcaption}
\usepackage{capt-of}
\usepackage{titlesec}

\usepackage{tabularx}
\usepackage{longtable}   % in preamble
\usepackage{tabularx}    % for flexible column widths
\usepackage{ltablex}     % allows longtable + tabularx

\usepackage{verbatim} % for commenting a block%

\title{Quantum-secure-by-construction (QSC): a paradigm shift for post-quantum agentic intelligence}

\author{
 Arit Kumar Bishwas\thanks{First and Corresponding author.} \\
  PricewaterhouseCoopers USA\\
  San Francisco, CA \\
  \texttt{arit.kumar.bishwas@pwc.com} \\
   \And
 Mousumi Sen\thanks{Co-first author.} \\
  Independent Researcher, USA\\
  Fremont, CA \\
  \texttt{mousumisen.official@gmail.com}\\
  \And
Albert Nieto-Morales\thanks{Second author.} \\
  PricewaterhouseCoopers USA\\
  San Francisco, CA \\
  \texttt{albert.morales@pwc.com} \\
\And
Joel Jacob Varghese\thanks{Third author.} \\
  PricewaterhouseCoopers USA\\
  San Francisco, CA\\
  \texttt{joel.jacob.varghese@pwc.com} \\
}

\usepackage{titlesec}              % controls section skips
\titlespacing*{\section}{0pt}{*3}{0.5pt} % after-skip = 0 pt

\begin{document}
\maketitle

%% Abstract
%%%%%%%%%%%%%%%%%%%%%%%%%%%%%%%%%%%%%%%%%%%%%%%%%%%%%%%%%%%%%%%%%%%%%%%%%%%%%%%%%%%%%%%%%%%%%%

\begin{abstract}
As agentic artificial intelligence systems scale across globally distributed and long lived infrastructures, secure and policy compliant communication becomes a fundamental systems challenge. This challenge grows more serious in the quantum era, where the cryptographic assumptions built into today’s AI deployments may not remain valid over their operational lifetime. Here, we introduce quantum secure by construction, or QSC, as a design paradigm that treats quantum secure communication as a core architectural property of agentic AI systems rather than an upgrade added later. We realize QSC through a runtime adaptive security model that combines post quantum cryptography, quantum random number generation, and quantum key distribution to secure interactions among autonomous agents operating across heterogeneous cloud, edge, and inter organizational environments. The approach is cryptographically pluggable and guided by policy, allowing the system to adjust its security posture according to infrastructure availability, regulatory constraints, and performance needs. QSC contributes a governance aware orchestration layer that selects and combines link specific cryptographic protections across the full agent lifecycle, including session bootstrap, inter agent coordination, tool invocation, and memory access. Through system level analysis and empirical evaluation, we examine the trade offs between classical and quantum secure mechanisms and show that QSC can reduce the operational complexity and cost of introducing quantum security into deployed agentic AI systems. These results position QSC as a foundational paradigm for post quantum agentic intelligence and establish a principled pathway for designing globally interoperable, resilient, and future ready intelligent systems.
\\ 

\end{abstract}

%% Section 1 Introduction
%%%%%%%%%%%%%%%%%%%%%%%%%

\section{Introduction}

The emergence of agentic artificial intelligence (AI) \cite{Ref01-AgenticAI-derouiche2025agentic} marks a fundamental transition in the design and deployment of intelligent systems. Moving beyond passive, task-specific models \cite{Ref02-GPT4-bubeck2023sparks} \cite{Ref03-anthropic2025claude4} \cite{Ref04-meta2025llama4}, agentic AI systems exhibit sustained reasoning, persistent memory, autonomous decision-making, real-world interaction, and dynamic tool orchestration. Once confined to isolated demonstrations or monolithic platforms, these systems are now scaling into globally distributed, cooperative ecosystems. Deployed across cloud infrastructures, edge devices, terrestrial networks, and satellite systems, agentic AI enables a new class of intelligence characterized by decentralized coordination, contextual adaptation, and long-lived autonomy across geographical and institutional boundaries. As agentic AI systems evolve toward globally distributed, autonomous ecosystems involving thousands of interacting agents, the security of inter-agent communication becomes a foundational systems problem. Ensuring that these systems remain secure across long operational lifecycles requires integrating cryptographic guarantees directly into the architecture of AI platforms rather than treating security as a retrofitted layer.

This architectural evolution introduces a qualitatively different security landscape. However, prevailing security approaches remain largely channel-centric (e.g., strengthening transport links) rather than lifecycle-centric (securing orchestration, tool-use, and persistent agent state end-to-end). PQC-only upgrades address computational attacks on public-key primitives, but do not by themselves provide deployment-time entropy hardening or information-theoretic key material where quantum channels exist. Conversely, QKD-only deployments are constrained by infrastructure availability and cannot serve as a uniform baseline across heterogeneous cloud/edge jurisdictions, creating a need for policy-driven cryptographic agility at runtime. As agentic systems operate with increasing autonomy across heterogeneous and often untrusted environments, secure, real-time, and decentralized communication becomes a foundational requirement rather than an implementation detail. Multi-agent coordination, ranging from collaborative planning and negotiation to autonomous execution, relies on continuous state synchronization, identity verification, and trust establishment over potentially adversarial networks. Existing threat vectors, including impersonation attacks, data poisoning, and cross-network session hijacking, already strain conventional AI security models. At global scale, these risks compound, directly affecting regulatory compliance, operational continuity, and trust guarantees across jurisdictions.

Concurrently, rapid advances in quantum computing are reshaping the long-term security assumptions underlying distributed digital systems. Quantum technologies are increasingly impacting diverse domains, including machine learning \cite{Ref4_1-Biswas2018QuantumSVM}\cite{Ref4_2-Biswas2020QuantumClustering}\cite{Ref4_3-quantumGauKernel_doi:10.1142/S0219749920500069}\cite{Ref4_4_QuantumNeuralNet_GarciaMartin2025QNNGP}\cite{Ref4_5_QuantumMachineLearning_Biamonte2017QML}, combinatorial optimization \cite{Ref4_9_Patent_Bishwas2024TrafficQuantum}\cite{Ref4_10_Patent_Choudhary2024EntityAllocation}, physical system simulation \cite{Ref4_8_Bishwas2024MolecularUnfolding}, blockchain technologies \cite{Ref4_6_QBlockChain_Reddy2025QuantumBlockchain}, and secure communication \cite{Ref010-QKD-Nature-zheng2025fs-cvqkd}. At the same time, cryptographically capable quantum computers pose a structural threat to widely deployed public-key cryptosystems \cite{Ref1-shor1994}\cite{Ref2-grover1996}\cite{Ref3-nistpqc}\cite{Ref4-bernstein2009}\cite{Ref5-mosca2018}. Once such systems mature, foundational primitives including RSA \cite{ref_sec1_RSA_Rivest1978RSA}, elliptic-curve cryptography \cite{ref_sec1_ECC_Wohlwend2016ECC}, and Diffie--Hellman key exchange \cite{ref_sec1_Diffie-Hellman_Merkle1978Secure} become vulnerable to quantum algorithms such as Shor’s \cite{Ref1-shor1994} and Grover’s \cite{Ref2-grover1996}. Moreover, encrypted communications transmitted today are susceptible to harvest-now-decrypt-later attacks \cite{Ref13-schneier1996}\cite{Ref25-liu2022}, in which adversaries archive data for future quantum decryption.

Agentic AI systems are particularly exposed to this emerging threat model. Their distributed topology, persistent state, and mission-critical intelligence amplify the consequences of long-term confidentiality and integrity failures. Unlike stateless services, agentic systems are often designed to operate continuously over extended lifetimes, making post-deployment cryptographic migration significantly more complex and costly. Retrofitting quantum-resistant security into already deployed agentic infrastructures risks disrupting agent coordination, violating compliance guarantees, and introducing new attack surfaces during transition. These characteristics motivate a shift from reactive cryptographic upgrades toward security models that are quantum-resilient by design.

Our contributions are threefold. First, we formalize QSC as a system-level principle for securing globally distributed agentic AI across heterogeneous trust boundaries. Second, we introduce a runtime policy-driven mechanism that selects link-specific cryptographic postures and derives unified session keys by composing PQC, QRNG, and (when available) QKD. Third, we empirically validate feasibility through micro-benchmarking and cloud-scale deployment analysis, including robustness under adversarial tamper and replay conditions. In this work, we advance \emph{quantum-secure-by-construction} (QSC) as a paradigm for post-quantum agentic intelligence, in which quantum-secure communication is embedded as a first-class property of agentic systems from inception rather than appended after deployment. We operationalize this paradigm through a composable security model that integrates three complementary cryptographic layers \cite{Ref26-bishwas2024}\cite{Ref05-manageCyberSec-AKB-9698591}\cite{Ref4_7_NietoMorales2025QuantumAES}:

\begin{itemize}
    \item \textbf{Post-quantum cryptography (PQC):} Lattice- and hash-based algorithms standardized by NIST to replace classical public-key schemes vulnerable to quantum attacks.
    \item \textbf{Quantum random number generation (QRNG):} Physically sourced entropy to eliminate predictability in key material and strengthen authentication and session establishment.
    \item \textbf{Quantum key distribution (QKD):} Information-theoretically secure key exchange for high-assurance communication links where quantum infrastructure is available.
\end{itemize}

This layered approach is cryptographically pluggable and policy-driven, enabling runtime adaptation to heterogeneous infrastructure capabilities, jurisdictional constraints, and performance requirements. By integrating quantum-secure primitives directly into agent identity, communication, and lifecycle management, the proposed paradigm supports globally interoperable, resilient, and forward-compatible agentic AI systems.

Taken together, these contributions position quantum-secure-by-construction as a foundational design principle for secured AI in the post-quantum era, bridging the domains of multi-agent intelligence and quantum cryptography and establishing a practical pathway toward long-lived, autonomous systems resilient to both present and future adversaries.

%% Section 2 Background
%%%%%%%%%%%%%%%%%%%%%%%
\section{Background}

\subsection{Agentic AI systems}

Modern agentic artificial intelligence (AI) systems \cite{Ref01-AgenticAI-derouiche2025agentic} increasingly adopt modular and composable designs in which autonomous agents collaborate across distributed environments to achieve complex objectives. Unlike monolithic AI models, agentic systems are composed of multiple interacting entities that exhibit persistent state, goal-directed behavior, and the ability to reason, communicate, and act both independently and collectively. As these systems scale, agents may be deployed across heterogeneous substrates, including cloud infrastructures, edge devices, robotic platforms, and satellite networks which operate under diverse latency, trust, and governance constraints.

At their core, multi-agent systems (MAS) consist of intelligent agents endowed with capabilities for perception, retrieval, memory, reasoning, communication, and action. Through perception and input interfaces, agents ingest multimodal signals originating from users, software systems, physical sensors, and peer agents, transforming heterogeneous observations into internal state representations that support downstream inference and decision-making. Retrieval-augmented generation (RAG) modules further extend agent capabilities by enabling access to external knowledge sources such as vector databases, enterprise repositories, and APIs, thereby grounding responses in current and domain-specific context rather than relying solely on parametric model knowledge \cite{Ref06-RAG-gao2023ragsurvey}. Memory and belief stores preserve both short-term interaction context and long-term knowledge, while planning and reasoning engines decompose goals, evaluate alternatives, and adapt execution in response to changing conditions \cite{Ref07-planningAndreasoning-sapkota2025agentic}. Tools, actuators, and communication interfaces then allow agents to invoke APIs, execute workflows, control physical systems, and coordinate with other agents or supervisory components \cite{Ref01-AgenticAI-derouiche2025agentic}. Collectively, these modules distinguish agentic AI from passive generative systems by enabling persistent, stateful, and goal-directed autonomy across dynamic environments.

As agentic AI systems transition from isolated deployments to globally distributed infrastructures, their architectural properties introduce new systemic requirements. Continuous inter-agent communication, shared memory access, and coordinated decision-making must occur across organizational and jurisdictional boundaries, often in the absence of centralized trust anchors. This shift elevates communication integrity, identity assurance, policy compliance, and long-term operational resilience from implementation concerns to foundational system properties. Consequently, agentic AI systems must be designed not only for autonomy and scalability, but also for verifiability, resilience, and sustained alignment with evolving security and regulatory constraints.

\subsection{Evolving multi-agent AI: from local to global}

As artificial intelligence progresses beyond single-agent operation, multi-agent systems (MAS) have become central to addressing complex, distributed problems. These systems comprise autonomous agents that coordinate through planning, memory, and communication to achieve shared or independent objectives. Historically, MAS were largely confined to localized deployments such as data centers, cloud regions, industrial facilities, or edge environments, where trust boundaries, latency, and infrastructure remained relatively controlled. Increasingly, however, MAS are expanding across geographically and administratively distinct domains, including multi-cloud infrastructures, cross-border enterprise networks, satellites, and mobile or edge platforms. This transition enables greater scale and capability, but also introduces stronger requirements for interoperability, jurisdictional compliance, identity assurance, and trusted communication across heterogeneous environments. As a result, globally distributed MAS must be designed not only for coordination and scalability, but also for secure and resilient operation in settings where secure, trustworthy inter-agent collaboration is a prerequisite for effective agentic AI.

\subsection{Quantum cryptographic framework}

\subsubsection{Layer 1: Post-quantum cryptography (PQC)}

Post-quantum cryptography (PQC) addresses the vulnerability of widely deployed public-key cryptosystems to quantum algorithms, providing quantum-resistant alternatives based on lattice- and hash-based hardness assumptions. As cryptographically capable quantum computers emerge, primitives such as RSA and elliptic-curve cryptography become insecure, motivating the adoption of PQC to ensure long-term confidentiality and authentication. Recent standardization efforts by NIST have accelerated the transition toward deployable quantum-resistant primitives.

Within agentic AI systems, PQC serves as the foundational security layer for communication and identity management. NIST-standardized algorithms, including Kyber \cite{Ref08-Kyber-cryptoeprint:2017/634} for key encapsulation and Dilithium \cite{Ref09-Dilithium-cryptoeprint:2017/633} for digital signatures, enable secure establishment of encrypted channels and authenticated message exchange across REST, RPC, and data-plane interactions. These primitives are well suited to the asynchronous, stateless communication patterns characteristic of large-scale, distributed agentic platforms, providing a practical and immediately deployable basis for quantum-resistant operation.

\subsubsection{Layer 2: Quantum random number generation (QRNG)}

Quantum random number generation (QRNG) derives entropy from intrinsically stochastic quantum processes, providing non-deterministic randomness that cannot be predicted or reproduced by classical algorithms \cite{Ref12-idq}. In contrast to pseudo-random number generators, which may suffer from entropy degradation or state compromise, QRNG offers high-integrity entropy suitable for cryptographic key generation, authentication tokens, and digital signatures.

Within agentic AI systems, QRNG strengthens trust at the entropy source by supplying uncorrelated randomness for agent identity initialization, session key establishment, and secure task delegation. QRNG can be integrated through embedded hardware devices for physical agents, such as drones or satellites, or accessed via remote quantum entropy services in cloud environments. This flexibility enables consistent entropy assurance across heterogeneous deployments, reinforcing quantum-secure-by-construction designs at the foundational level.

\subsubsection{Layer 3: Quantum key distribution (QKD)}

Quantum key distribution (QKD) enables two parties to establish shared encryption keys with information-theoretic security by exploiting fundamental quantum principles, including the no-cloning theorem and measurement-induced disturbance \cite{Ref12-idq, Ref010-QKD-Nature-zheng2025fs-cvqkd}. Unlike classical key exchange protocols that rely on computational hardness assumptions, QKD provides intrinsic eavesdropping detection, rendering passive interception ineffective even in the presence of cryptographically capable quantum adversaries.

Within quantum-secure-by-construction agentic systems, QKD serves as a high-assurance communication layer where quantum infrastructure is available. Such links are particularly relevant for security-critical channels, including satellite–ground communication, inter–data center connectivity across jurisdictions, and coordination among high-trust organizational domains. In adversarial or untrusted network environments, measurement-device-independent QKD (MDI-QKD) further mitigates implementation-level vulnerabilities, reinforcing end-to-end key establishment guarantees.

%% Section 3 Proposed Quantum-Secure Agentic AI Architecture
%%%%%%%%%%%%%%%%%%%%%%%%%%%%%%%%%%%%%%%%%%%%%%%%%%%%%%%%%%%%

\section{Quantum-Secure-by-Construction Agentic AI System Model}

As agentic AI systems are increasingly deployed to autonomously solve complex, distributed problems, trust and security across interacting agents emerge as foundational system requirements. Classical cryptographic mechanisms provide limited long-term assurance in this setting, as their security assumptions are fundamentally challenged by the advent of cryptographically capable quantum computers. For long-lived, globally distributed agentic systems, post-deployment cryptographic migration introduces significant operational complexity and risk, motivating security models that anticipate quantum-era threats from inception.

In this work, we instantiate the quantum-secure-by-construction (QSC) paradigm through a system-level model for globally distributed agentic AI. Unlike prior quantum-enhanced agentic AI studies that focus on isolated quantum communication channels or localized algorithmic acceleration \cite{Ref29-Jenefa2024,Ref31-Acha2025,Ref30-Zhao2022,Ref32-Akter2023}, the proposed model provides an end-to-end, runtime-adaptive cryptographic foundation integrated across the agent lifecycle. The system is organized into three interacting operational layers—a global coordination layer, a task agent layer, and a quantum cryptographic layer—each reinforced through the coordinated application of post-quantum cryptography (PQC), quantum random number generation (QRNG), and quantum key distribution (QKD) \cite{Ref6-alkim2016,Ref7-peikert2016,Ref8-buchmann2009,Ref9-lucamarini2018,Ref10-ma2016,Ref11-toshiba2022,Ref12-idq,Ref010-QKD-Nature-zheng2025fs-cvqkd}.

By embedding the quantum cryptographic stack across all communication links and agent-internal pathways, cryptographic protection is applied selectively according to threat exposure, infrastructure availability, and operational criticality. This approach transforms a distributed collection of autonomous agents—each equipped with reasoning, memory, retrieval, and tool-invocation capabilities—into a verifiable and quantum-resilient agentic ecosystem capable of operating across heterogeneous jurisdictions, volatile networks, and adversarial environments.

\begin{figure}[h!]
    \centering
    \includegraphics[width=0.9\textwidth]{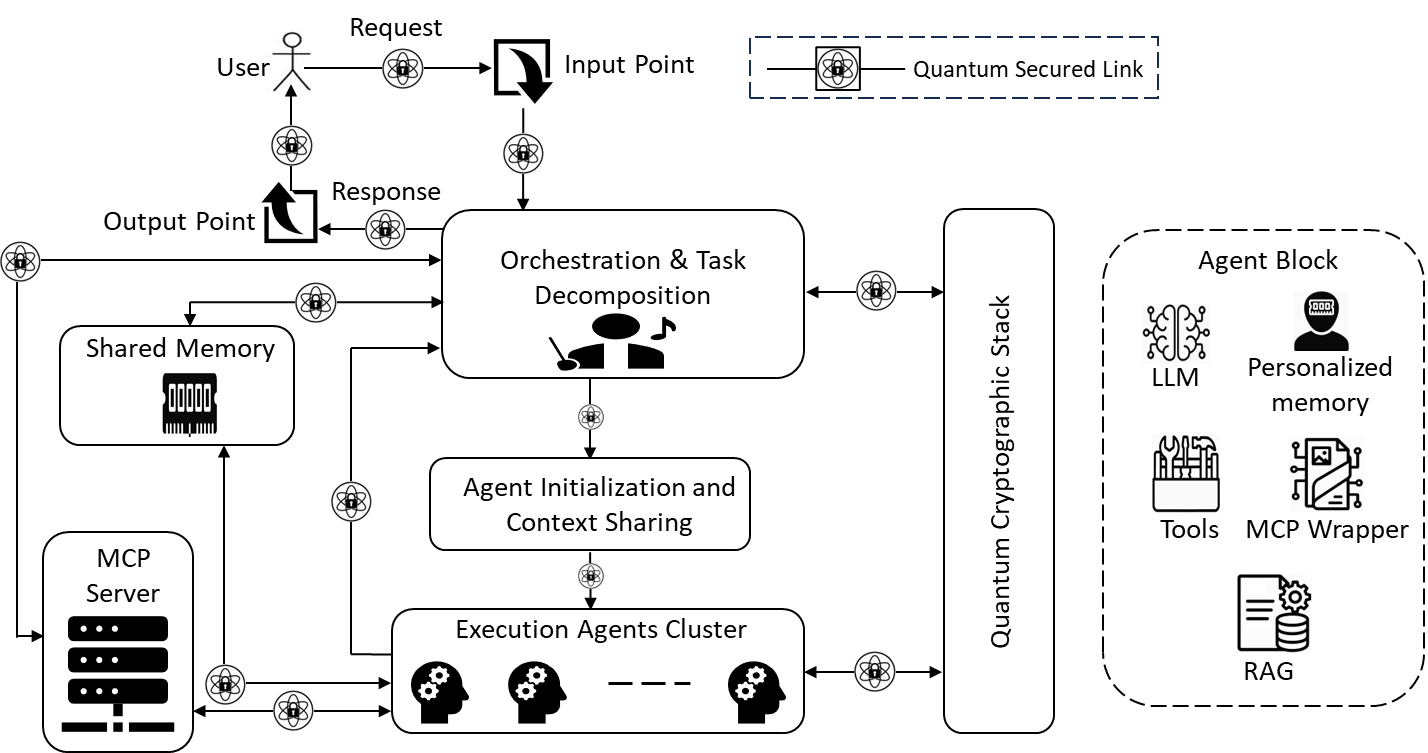}
    \caption{Quantum-secure-by-construction agentic AI system model.}
    \label{fig:Figure1}
\end{figure}

Figure~\ref{fig:Figure1} illustrates the proposed system model for a globally distributed, quantum-secure-by-construction agentic AI deployment. Each agent—whether operating as a global coordinator or as a task-specialized entity—functions as a self-contained cognitive system with modular perception, reasoning, memory, retrieval, and action components. All inter-agent and intra-agent communication pathways are secured using quantum-resilient cryptographic mechanisms, ensuring that operations such as memory access, retrieval augmentation, tool invocation, and policy enforcement remain protected against both classical and quantum adversaries throughout the agent lifecycle.

\subsection{Formal system architecture}
\label{sec:formal_architecture}

To enable rigorous analysis of orchestration, execution, and security properties under the quantum-secure-by-construction (QSC) paradigm, we formalize the agentic AI system as a directed interaction graph. The system is represented as
\begin{equation}
\mathcal{S} = (\mathcal{V}, \mathcal{E}),
\label{eq:system_graph}
\end{equation}
where the vertex set
\begin{equation}
\mathcal{V} = \{ C, O \} \cup \mathcal{A}
\label{eq:vertex_set}
\end{equation}
comprises a client $C$, a global orchestrator $O$, and a collection of autonomous agents
$\mathcal{A} = \{ A_1, A_2, \dots, A_n \}$. Each directed edge $(u,v) \in \mathcal{E}$
denotes a communication, coordination, or control channel through which information,
commands, or state are exchanged.

For a given client request, the orchestrator dynamically constructs a directed acyclic
task graph (DAG)
\begin{equation}
G = (V, E),
\label{eq:task_dag}
\end{equation}
where each vertex $\tau_i \in V$ represents an atomic task and each directed edge
$(\tau_i, \tau_j) \in E$ encodes an execution dependency. Task execution is determined
by an assignment function
\begin{equation}
\rho : V \rightarrow \mathcal{A},
\label{eq:task_assignment}
\end{equation}
which maps each task to the agent responsible for its execution.

This formal representation establishes a direct correspondence between conceptual
agentic system components and mathematical objects, enabling precise reasoning about
task flow, inter-agent coordination, and the placement of cryptographic controls.
In particular, modeling communication pathways as explicit graph edges provides a
natural foundation for enforcing adaptive, link-specific security guarantees under
the QSC paradigm.

\subsection{Cryptographic posture and policy selection}
\label{sec:crypto_posture}

Under the quantum-secure-by-construction (QSC) paradigm, security enforcement is
realized through adaptive cryptographic postures applied to individual communication
links within the agentic system. Let $\mathcal{E}$ denote the set of directed edges
in the system graph $\mathcal{S} = (\mathcal{V}, \mathcal{E})$, where each edge
$(u,v) \in \mathcal{E}$ represents a logical communication channel from entity $u$
to entity $v$.

For each such link, the system assigns a cryptographic posture
\begin{equation}
\pi_{u,v} \in \Pi = \{\textsf{PQC},\ \textsf{PQC{+}QRNG},\ \textsf{QKD{+}PQC{+}QRNG}\},
\label{eq:crypto_posture}
\end{equation}
where \textsf{PQC} denotes exclusive use of post-quantum public-key primitives,
\textsf{PQC{+}QRNG} augments post-quantum cryptography with quantum-derived entropy,
and \textsf{QKD{+}PQC{+}QRNG} combines quantum key distribution with post-quantum
cryptography and quantum random number generation for high-assurance links.

Cryptographic posture selection is governed by a dynamic security policy agent,
which evaluates security guarantees against operational constraints at runtime.
Formally, the selected posture for link $(u,v)$ is obtained by solving
\begin{equation}
\pi_{u,v} =
\arg\max_{\pi \in \Pi}
\Big(
\lambda_s \cdot \mathrm{Sec}(\pi; u, v)
- \lambda_\ell \cdot \mathrm{Lat}(\pi; u, v)
- \lambda_c \cdot \mathrm{Cost}(\pi; u, v)
\Big),
\label{eq:posture_optimization}
\end{equation}
subject to the compliance constraint
\begin{equation}
\mathrm{Comp}(\pi; \mathrm{region}(u), \mathrm{region}(v)) = 1.
\label{eq:compliance_constraint}
\end{equation}

Here, $\mathrm{Sec}(\pi; u, v)$ quantifies the cryptographic security strength provided by posture $\pi$ on link $(u,v)$, $\mathrm{Lat}(\pi; u, v)$ denotes the associated latency overhead, and $\mathrm{Cost}(\pi; u, v)$ captures computational and infrastructure costs. The non-negative weighting coefficients $\lambda_s$, $\lambda_\ell$, and $\lambda_c$ encode system-level priorities over security, performance, and cost, respectively.

The compliance function $\mathrm{Comp}(\cdot)$ enforces regulatory, jurisdictional, and infrastructural constraints associated with the geographic regions in which entities $u$ and $v$ operate. Operationally, the policy agent instantiates Sec(·), Lat(·), and Cost(·) using mission criticality, peer trust state, QKD link availability, regional compliance requirements, and latency budgets, enabling graceful escalation to QKD assisted postures when available and compliant. This formulation enables link-specific, policy-driven security adaptation across heterogeneous environments, providing cryptographic agility while preserving regulatory compliance and operational efficiency.

\subsection{Unified session key derivation}
\label{sec:unified_key_derivation}

To ensure cryptographic continuity across heterogeneous deployment environments, the quantum-secure-by-construction (QSC) paradigm employs a unified session key derivation mechanism that composes post-quantum, quantum-random, and quantum-distributed entropy sources. This construction enables consistent security semantics across both quantum-enabled and quantum-constrained settings.

Let $(u,v) \in \mathcal{E}$ denote a directed communication link between entities $u$ and $v$. For each such link, the following keying materials are defined:
\begin{itemize}
  \item $K^{\mathrm{qkd}}_{u,v}$: symmetric key material generated via quantum key
  distribution (QKD) when a quantum channel between $u$ and $v$ is available;
  otherwise $K^{\mathrm{qkd}}_{u,v} = \bot$.
  \item $K^{\mathrm{pqc}}_{u,v}$: a shared secret derived from a post-quantum key
  encapsulation mechanism (KEM) executed over classical communication channels.
  \item $r_{u,v}$: fresh entropy sampled from a quantum random number generator
  (QRNG), providing session-level freshness and entropy diversification.
\end{itemize}

The session key used to protect communication over link $(u,v)$ is derived as
\begin{equation}
K_{u,v} =
\textsf{HKDF}\Big(
\mathbf{1}[K^{\mathrm{qkd}}_{u,v} \neq \bot]\cdot K^{\mathrm{qkd}}_{u,v}
\ \Vert\ 
K^{\mathrm{pqc}}_{u,v}
\ \Vert\ 
r_{u,v}
\Big),
\label{eq:unified_key_derivation}
\end{equation}
where $\textsf{HKDF}(\cdot)$ denotes a cryptographic hash-based key derivation
function, $\Vert$ denotes concatenation, and $\mathbf{1}[\cdot]$ is the indicator
function.

This formulation ensures that QKD-derived entropy contributes directly to the
session key whenever quantum channels are available, while preserving
post-quantum security guarantees through $K^{\mathrm{pqc}}_{u,v}$ in purely
classical environments. The inclusion of $r_{u,v}$ enforces cryptographic freshness and mitigates key reuse across sessions, yielding a unified key establishment mechanism that remains secure and interoperable across heterogeneous agentic deployments.

\subsection{Secure message exchange protocol}
\label{sec:secure_message_exchange}

Communication between entities in a quantum-secure-by-construction (QSC) agentic system follows a secure message exchange protocol that provides confidentiality, integrity, authenticity, and replay protection at the link level.

Let $(u,v) \in \mathcal{E}$ denote a directed communication link, and let $K_{u,v}$
be the session key derived as described in
Section~\ref{sec:unified_key_derivation}. For a message payload $m$ transmitted
from entity $u$ to entity $v$, the sender samples a fresh nonce
\begin{equation}
n \leftarrow \textsf{QRNG},
\label{eq:nonce_generation}
\end{equation}
where uniqueness is enforced per session and message instance.

Authenticated encryption is applied to the payload as
\begin{equation}
c = \textsf{AEAD.Enc}(K_{u,v}, n, m; \mathrm{aad}),
\label{eq:aead_encryption}
\end{equation}
where $\textsf{AEAD.Enc}(\cdot)$ denotes an authenticated encryption scheme and
$\mathrm{aad}$ represents associated authenticated data, including protocol
metadata and contextual identifiers.

To ensure message authenticity and non-repudiation, the sender computes a digital
signature
\begin{equation}
\sigma = \textsf{Sig.Sign}\big(\mathsf{sk}_u, H(c \Vert \mathrm{aad} \Vert n)\big),
\label{eq:digital_signature}
\end{equation}
where $\mathsf{sk}_u$ denotes the private signing key of entity $u$ and $H(\cdot)$
is a cryptographic hash function.

The transmitted message takes the form
\begin{equation}
\mathsf{msg}_{u \rightarrow v} =
\langle u, v, n, \mathrm{aad}, c, \sigma \rangle.
\label{eq:message_format}
\end{equation}

Upon receipt, entity $v$ verifies the signature using the corresponding public key
$\mathsf{pk}_u$, decrypts the ciphertext using $K_{u,v}$, and enforces replay
protection by maintaining a set $\mathcal{N}_v$ of previously observed nonces. Any
message for which $n \in \mathcal{N}_v$ is rejected.

\subsection{Global coordination layer}

The global coordination layer acts as the system-level control plane for
distributed agentic AI, managing task orchestration, agent provisioning, and
policy enforcement across heterogeneous regions. As it mediates system-wide
decisions and trust relationships, this layer operates under the strictest
security assumptions of the quantum-secure-by-construction (QSC) paradigm. Post-quantum cryptography (PQC) secures orchestration, agent bootstrapping, and authenticated control signaling by default, while quantum random number
generation (QRNG) provides high-entropy, single-use session tokens to prevent
replay and spoofing. Quantum key distribution (QKD) is selectively applied to
protect control links spanning sovereign or high-assurance domains, ensuring
robustness against both classical and quantum adversaries. Designed to be failure-intolerant, this layer assigns tasks based on capability, location, and trust constraints; aggregates distributed results; and enforces
cryptographic posture and operational policies at runtime. PQC-signed commands and tamper-resistant coordination logs support auditability and accountability, maintaining secure execution and cryptographic resilience at scale.

\subsection{Task agent layer}

The task agent layer comprises autonomous agents responsible for executing assigned tasks and coordinating with peers and the global orchestrator. Each agent functions as a self-contained cognitive unit that integrates perception, retrieval-augmented generation (RAG), memory, planning and reasoning, tool invocation, and secure communication within a common execution boundary. Through perception and input interfaces, agents ingest multimodal data including text, audio, visual streams, sensor telemetry, external system feeds, and human interactions, and transform these heterogeneous signals into normalized internal representations for downstream decision-making. Retrieval-augmented generation (RAG) extends reasoning beyond static model parameters by incorporating relevant context from private vector stores, enterprise knowledge bases, external search interfaces, APIs, and long-term memory. Memory and belief stores maintain both short-term contextual coherence and long-term knowledge retention, supporting continuity and stateful adaptation across tasks. Planning and reasoning engines translate high-level objectives into executable plans while reasoning over dependencies, constraints, and coordination requirements, including symbolic, neural, and neuro-symbolic approaches \cite{Ref07-planningAndreasoning-sapkota2025agentic}. Tools and actuator interfaces allow agents to invoke APIs, execute code, interact with enterprise platforms, and control robotic or cyber-physical systems under policy constraints \cite{Ref01-AgenticAI-derouiche2025agentic}. Secure communication interfaces support both real-time and asynchronous interaction among agents, orchestrators, and external services, while the Model Context Protocol (MCP) provides an emerging coordination substrate for context propagation, tool invocation, and role-aware interaction across distributed agentic systems \cite{Ref06-RAG-gao2023ragsurvey,Ref012-MCP-hou2025mcp,ref_sec3.2.6_CrewAI_Website,ref_sec3.2.6_MicrosoftAutoGenDocs}.

Under the quantum-secure-by-construction (QSC) paradigm, all intra-agent and inter-module paths are protected through a unified cryptographic posture rather than separate security treatments at each functional block. Post-quantum cryptography (PQC) secures internal and external exchanges, with Kyber \cite{Ref08-Kyber-cryptoeprint:2017/634} for encryption and Dilithium \cite{Ref09-Dilithium-cryptoeprint:2017/633} or SPHINCS+ \cite{ref_sec3.2.7_SPHINCS_Website} for authentication and integrity, thereby preserving confidentiality, authenticity, and tamper resistance against both classical and quantum adversaries. Quantum random number generation (QRNG) supplies high-entropy keys, nonces, freshness material, and replay-resistant tokens across perception, memory, planning, tool invocation, and communication flows \cite{Ref27-Cao2015}. For a subset of high-assurance deployments, including defense, sovereign, and space-ground environments, quantum key distribution (QKD) may be selectively applied to establish information-theoretically secure symmetric keys for especially sensitive channels. Together, these mechanisms ensure that all intra-agent operations, memory accesses, reasoning artifacts, tool invocations, and communication links remain secure, policy-compliant, and verifiable across heterogeneous deployment environments.

\subsection{Quantum cryptographic stack}

The quantum cryptographic stack constitutes the security substrate of the agentic
AI system, protecting communication, computation, and persistent state against
quantum-capable adversaries. It may be deployed as an embedded capability within
individual agents or exposed as a shared service through quantum communication
infrastructure, including QKD fiber networks \cite{Ref28-Pirandola2019}, satellite
links, or cloud-based quantum entropy services, enabling consistent operation
across edge, cloud, and space-based environments.

Security enforcement follows a tiered fallback model. Where available, quantum
key distribution (QKD) provides information-theoretic protection for high-assurance
links. In the absence of quantum channels, post-quantum cryptography (PQC),
including Kyber and Dilithium, serves as the default mechanism for confidentiality
and authentication. Quantum random number generation (QRNG) supplies high-entropy
material for keys and nonces, with monitored fallback to trusted pseudo-random
sources under degraded conditions. This layered design ensures continuous,
quantum-resilient protection even under partial infrastructure availability.

A dynamic quantum security policy agent (DQSPA) supervises the cryptographic stack,
enabling adaptive security across the agent lifecycle. The DQSPA monitors advances
in post-quantum cryptography, quantum protocols, and regulatory requirements, and
automatically updates system configuration through key rotation, algorithm
upgrades, and policy enforcement. This self-adaptive mechanism maintains long-term
cryptographic resilience across heterogeneous and geographically distributed
agentic AI deployments, establishing security as a foundational system property
rather than a post-deployment augmentation.

%%%%%% Section 4 Operational Steps of a quantum-secure agentic AI system 
%%%%%%%%%%%%%%%%%%%%%%%%%%%%%%%%%%%%%%%%%%%%%%%%%%%%%%%

\section{Operational Steps of a Quantum-secure Agentic AI System}

This section describes the operational lifecycle of a globally distributed, quantum-secure-by-construction agentic AI system, from session initialization and task orchestration to inter-agent coordination and audit logging. Each stage integrates post-quantum cryptographic protection, quantum-derived entropy, and zero-trust principles to ensure security across the full execution pipeline. The operational workflow follows a modular, layered structure corresponding to the global coordination layer, task agent layer, and quantum cryptographic stack introduced earlier. Together, these steps constitute the executable realization of the proposed system model, enabling secure and policy-compliant operation across heterogeneous and geographically distributed environments.

\textbf{Step 1: Quantum-secured session bootstrap} 

Each client interaction begins with the establishment of a quantum-secured session. The global orchestrator initializes post-quantum cryptographic credentials using Kyber for key exchange and Dilithium for digital signatures, while session identifiers and nonces are generated via quantum random number generation (QRNG). Where quantum infrastructure is available, quantum key distribution (QKD) may additionally establish symmetric session keys.

\textbf{Step 2: Secure client request intake}

Client requests are received through a validated intake process that verifies digital signatures, enforces sanitization, and attaches QRNG-derived request identifiers. Transport security is provided via post-quantum TLS or symmetric encryption, such as AES-GCM with Kyber-established keys \cite{ref_AES_NIST2001AES}, ensuring that all downstream actions remain cryptographically bound to an authenticated request origin.

\textbf{Step 3: Task graph construction}

The orchestrator constructs a directed task graph following a LangGraph-style execution model \cite{ref_LangGraphDocs,ref_LangChain_LangChainIntro}, in which nodes represent specialized agents and edges capture coordination and data-flow dependencies expressed in Model Context Protocol (MCP) format \cite{Ref012-MCP-hou2025mcp}. To preserve integrity and provenance, graph components are bound to the session context through post-quantum signatures and QRNG-seeded identifiers, with additional verification details provided in Appendix~A.

\textbf{Step 4: Secure agent execution}

Each agent executes its assigned subgraph within an isolated zero-trust context across cloud, edge, or hybrid environments. Retrieval, memory access, planning, tool invocation, and inter-agent communication operate under quantum-secure controls, including post-quantum encryption, per-invocation QRNG-derived freshness, and optional hardware-backed attestation for edge deployments.

\textbf{Step 5: Merge and reason}

After task execution, the orchestrator invokes the reasoning engine to aggregate agent outputs into a coherent system-level response. All outputs are verified using post-quantum digital signatures, and the reasoning process preserves consistency with the execution graph \cite{ref_LangGraphDocs,ref_LangChain_LangChainIntro} while maintaining integrity and traceability under distributed execution.

\textbf{Step 6: Finalize and respond}

The finalized output is assembled according to the original Model Context Protocol (MCP) contract and returned to the client. The response is cryptographically sealed using post-quantum-secured symmetric encryption, such as AES-GCM with Kyber-established keys \cite{ref_AES_NIST2001AES}, and signed by the orchestrator to provide confidentiality, integrity, non-repudiation, and end-to-end traceability.

\textbf{Step 7: Tamper-evident audit trail}
\label{sec:tamper_evident_audit}

All security-critical and execution-related events are recorded in an immutable, cryptographically verifiable audit ledger to support governance, compliance, and forensic analysis. Records are authenticated with post-quantum digital signatures, linked through tamper-evident cryptographic commitments, and may be regionally replicated using QKD when available, with formal construction details deferred to Appendix~A.

\textbf{Step 8: Session teardown and zero-knowledge proof}

At completion of execution, the orchestrator securely terminates the session by invalidating or erasing ephemeral state, including temporary memory, cryptographic keys, and session tokens. Where required, the system may generate a zero-knowledge proof (ZKP) attesting that execution adhered to authorized policies and that no unauthorized data access or leakage occurred.

Refer to Appendix B (Table~\ref{tab:Table3}), where we have summarized the mapping of the quantum-secure-by-construction (QSC) agentic AI system to its operational steps, architectural layers, agent roles, and quantum security controls. This alignment illustrates how orchestration, agent execution, and cryptographic enforcement remain integrated throughout the system lifecycle, while detailed algorithmic specifications are provided in Appendix~A (Algorithms~1--9).

The proposed framework is domain-agnostic and applicable across a wide range of real-world settings, including decentralized finance, global healthcare, critical infrastructure, and space-based autonomy. Its layered integration of post-quantum cryptography, quantum randomness, and selective quantum key distribution enables secure and resilient multi-agent collaboration under adversarial and regulatory constraints. Representative domain-specific use cases are summarized in Appendix~B (Table~\ref{tab:use_cases}).

%% Section 5 eployment regimes and lifecycle implications
%%%%%%%%%%%%%%%%%%%%%%%%%%%%%%%%%%%%%%%%%%%%%%%%%%%%%%%%%

\section{Deployment Regimes and Lifecycle Implications}

The security properties of agentic AI systems are shaped not only by cryptographic
primitives, but also by how those primitives are embedded across deployment
regimes and system lifecycles. In this section, we examine how the
quantum-secure-by-construction (QSC) paradigm manifests across local, global, and
long-lived agentic systems, and why early integration of quantum-resilient security
fundamentally alters feasibility, cost, and long-term risk.

\subsection{Local and cloud-native agentic systems}

In local or cloud-native deployments, such as single-region data centers,
enterprise clouds, or controlled edge clusters, agentic AI systems typically
operate within relatively stable trust boundaries and network conditions. In
these environments, the incremental overhead of adopting QSC is limited.
Post-quantum cryptography can be introduced alongside existing transport
mechanisms, quantum-derived entropy can complement classical randomness sources,
and the use of quantum-secured links can be selectively confined to sensitive
control-plane communication.

When QSC is incorporated at system inception, cryptographic controls align
naturally with agent lifecycles, orchestration logic, and policy enforcement.
Agent identities, memory, task graphs, and audit trails are natively bound to
quantum-resilient primitives, eliminating the need for later architectural
refactoring. As a result, early-stage agentic systems can absorb quantum security
with minimal operational disruption while remaining compatible with evolving
threat models and regulatory requirements.

\subsection{Geo-distributed and sovereign agentic systems}

As agentic AI systems expand across geographic regions, organizational boundaries,
and legal jurisdictions, their security assumptions change substantially.
Communication paths traverse heterogeneous networks, data residency requirements
diverge, and trust must be established across independently governed
infrastructures. In such environments, retrofitting quantum security becomes
increasingly impractical.

Introducing QSC after deployment would require re-keying long-lived agents,
re-signing historical task graphs, re-encrypting distributed memory stores, and
re-establishing trust anchors across sovereign domains, often under conflicting
regulatory constraints. Partial upgrades further create asymmetric security
postures, exposing weakest-link vulnerabilities across the agent mesh. By contrast, systems designed with QSC from the outset treat cryptographic agility as a foundational property. Cryptographic postures can be dynamically selected,
jurisdiction-aware, and uniformly enforced across orchestration, execution, and
audit layers. This enables global agentic systems spanning cloud, edge, and
satellite infrastructure to operate securely without disruptive transitions,
making quantum resilience an inherent structural characteristic rather than a
deployment afterthought.

\subsection{Long-lived agentic systems and cryptographic irreversibility}

The strongest motivation for QSC arises in long-lived agentic systems that are
expected to operate for years or decades in domains such as finance, healthcare,
critical infrastructure, defense, and autonomous governance. These systems
accumulate state, memory, learned behavior, and historical audit records that
cannot be safely reprocessed once cryptographic foundations are compromised.

In such settings, cryptographic decisions become effectively irreversible. Data
encrypted today may be harvested for future decryption, identities established
under classical assumptions may become forgeable, and audit trails may lose
evidentiary value when signature schemes fail. Retrofitting quantum security after
the emergence of cryptographically capable quantum adversaries is therefore a
reactive mitigation rather than a preventative solution. Quantum-secure-by-construction reframes this challenge by aligning agentic system
design with cryptographic longevity. By embedding post-quantum primitives,
quantum-derived entropy, and quantum-secured key establishment directly into the
agent lifecycle, QSC ensures that trust does not decay over time. In this sense,
QSC is not merely a security enhancement, but a necessary precondition for the safe
deployment of autonomous, long-lived intelligent systems in the post-quantum era.

%% Section 6 Interoperability and global trust fabric
%%%%%%%%%%%%%%%%%%%%%%%%%%%%%%%%%%%%%%%%%%%%%%%%%%%%%

\section{Interoperability and Global Trust Fabric}

As agentic AI systems extend across sovereign jurisdictions, heterogeneous
infrastructures, and divergent regulatory regimes, a unified trust fabric becomes
essential. Such a fabric must support seamless interoperability while preserving
cryptographic guarantees, agent integrity, and operational responsiveness.
Interoperability challenges arise from conflicting cryptographic policies,
heterogeneous hardware capabilities, and uneven availability of quantum
infrastructure. Table~\ref{tab:Table4_interoperability-matrix} summarizes representative interoperability regimes across regions and cryptographic layers, rather than providing an exhaustive account of legal requirements.

%\addtocounter{table}{-1}
\begin{table}[H]
\centering
\caption{Interoperability regimes across jurisdictions and quantum security layers}
%\renewcommand{\arraystretch}{1.3}
%\begin{tabularx}{\textwidth}{|X|X|X|X|X|}
\begin{tabularx}{\textwidth}{XXXXX}
\hline
\textbf{Region / Jurisdiction} & \textbf{Layer 1: PQC (Kyber, Dilithium)} & \textbf{Layer 2: QRNG} & \textbf{Layer 3: QKD} & \textbf{Compliance Notes} \\
\hline
\hline
United States (FIPS/NIST \cite{ref_sec6.1_FIPS-140-3_NISTFIPS1403}) & Full support (NIST standard) & Supported in national clouds and DoD networks & Restricted to defense/research use & FIPS-140 compliance required for operational systems \\
\hline
European Union (ENISA/GDPR) \cite{ref_sec6.1_ENISA2021EECC} & Fully supported & Available in some member states & Limited by cross-border data sovereignty laws & GDPR \& eIDAS integration; ENISA PQC draft alignment \\
\hline
China (Commercial Encryption Law) \cite{ref_sec6_Table_China_LaskaiSegal2021ChinaEncryption} & Partial (non-NIST PQC under review) & National QRNG standards vary & Strong QKD support (Beijing \(\leftrightarrow\) Shanghai) & Local encryption vendors must be certified \\
\hline
Japan & Fully supported & R\&D deployment in Tokyo cloud centers & Emerging satellite-QKD projects & METI and MIC jointly defining crypto standards \cite{ref_sec6_Table_METI2024AIGuidelines} \cite{ref_sec6_Table_Japan_Lightspark2025CryptoJapan} \\
\hline
Australia & Fully supported & Limited QRNG deployment & Under review for GovCloud and critical infrastructure & ASD-approved PQC needed for national defense AI \cite{ref_sec6_Table_Australia_ASD2022PQC} \\
\hline
Hybrid / Multinational Deployments & Mandatory baseline for all agents & Conditional per-node availability & Optional—negotiated QKD tunnels per jurisdiction & Crypto-pluggable fallback to PQC-only mode \\
\hline
\end{tabularx}
\label{tab:Table4_interoperability-matrix}
\end{table}

\subsection{Cross-jurisdiction cryptographic compliance}

To enable lawful and secure interoperation across jurisdictions, the proposed
framework adopts a layered cryptographic model in which post-quantum cryptography
(PQC) serves as the universal security baseline. Standardized PQC primitives,
including Kyber for key encapsulation and Dilithium for digital signatures, ensure
consistent agent authentication, secure communication, and identity validation
while aligning with emerging global standards such as NIST PQC \cite{Ref3-nistpqc},
FIPS~140-3 \cite{ref_sec6.1_FIPS-140-3_NISTFIPS1403}, and ENISA guidance
\cite{ref_sec6.1_ENISA2021EECC}.

Where permitted by regulatory and hardware constraints, quantum random number
generation (QRNG) augments PQC by providing high-assurance entropy for session keys,
nonces, and identity tokens. Quantum key distribution (QKD), as the most
infrastructure-dependent layer, is selectively enabled in regions with compliant
optical or satellite-based capabilities, such as EuroQCI in Europe
\cite{ref_sec6.1_EuroQCI_EC} or satellite QKD deployments in China
\cite{ref_sec6.1_China_Micius_satellite_Yin2020Entanglement}.

The architecture is crypto-pluggable by design. Agents dynamically adjust their
cryptographic posture based on geographic location, regulatory context, and peer
capabilities, falling back to PQC-only operation where quantum infrastructure is
unavailable, and enabling hybrid modes when partial quantum resources are present,
without requiring system-level reconfiguration.

\subsection{Governance, cryptographic orchestration, and adaptability}

A global agentic AI fabric requires governance-aware cryptographic orchestration rather than static security configurations. The orchestrator, operating in centralized or federated form, governs access control, cryptographic channel selection, auditability, and policy enforcement across the system. By maintaining a real-time view of agent trust states, cryptographic posture, and regional capabilities, it can dynamically initiate QKD sessions, trigger PQC key rotation, and reroute sensitive communication through trusted intermediaries. Continuous monitoring of agent behavior and cryptographic health is therefore essential. Metrics such as entropy quality, credential freshness, and handshake reliability help prevent cryptographic drift, replay vulnerabilities, and stale credentials, particularly in asynchronous or intermittently connected deployments such as space--Earth missions \cite{ref_sec6.2_Space_Earth_missions_NASA_ESTO}.

To remain viable under rapid advances in quantum technology and shifting regulatory environments, this governance layer must also support adaptability at the protocol level. Agents periodically assess upgrade readiness for emerging post-quantum schemes, transport mechanisms, and fallback pathways, while cryptographic fallback logic remains extensible enough to incorporate alternative entropy sources or distributed validation methods if centralized infrastructure becomes unavailable or compromised. In this way, the trust fabric can preserve secure and trusted interaction across heterogeneous geopolitical and operational settings, remaining robust, quantum-resilient, and forward compatible by design.

\subsection{Regulatory-aware cryptographic layering}

Agentic systems operating across national boundaries must satisfy diverse
cryptographic regulations and infrastructure constraints. To address this, the
proposed security stack is structured into three modular layers. Post-quantum
cryptography (PQC) serves as the universal compliance anchor, employing
NIST-standardized schemes such as Kyber and Dilithium to enable secure
authentication and communication under regimes governed by FIPS (U.S.), ENISA
(EU), and other sovereign cryptographic frameworks.

Quantum random number generation (QRNG) and quantum key distribution (QKD) are
selectively enabled where legally permitted and infrastructurally supported, such
as in national research clouds or defense-grade networks. In jurisdictions where
quantum capabilities are constrained or prohibited, agents automatically revert
to PQC-only operation while preserving cryptographic integrity and compliance.

This layered, crypto-pluggable design ensures legal interoperability while
maximizing security where possible. The architecture supports dynamic
reconfiguration, allowing quantum-enhanced layers to be activated or withdrawn at
runtime without disrupting system operation or violating regulatory constraints.

\subsection{Performance and overhead analysis}
\label{sec:performance_overhead}

While the primary objective of the proposed framework is long-term,
quantum-resilient security, it is also important to characterize the performance
overhead introduced by layered cryptographic enforcement. We present a simplified
analytical model to reason about latency and scalability trade-offs in distributed
agentic execution.

Let $\mathcal{E}$ denote the set of directed communication links used during task
graph execution. For each link $(u,v) \in \mathcal{E}$, communication latency is
decomposed into a one-time session setup cost and a per-message transmission cost. The total end-to-end communication time is approximated as
\begin{equation}
T_{\mathrm{total}} \approx
\sum_{(u,v) \in \mathcal{E}}
\Big(
T^{\pi_{u,v}}_{\mathrm{handshake}}
+
N_{u,v} \cdot T^{\pi_{u,v}}_{\mathrm{msg}}
\Big),
\label{eq:total_latency}
\end{equation} where $T^{\pi_{u,v}}_{\mathrm{handshake}}$ denotes the session establishment latency
under cryptographic posture $\pi_{u,v}$, $T^{\pi_{u,v}}_{\mathrm{msg}}$ represents
the average per-message processing and transmission latency, and $N_{u,v}$ is the
number of messages exchanged over the link. This formulation makes explicit the trade-offs among cryptographic postures.
QKD-enabled links typically incur higher handshake latency due to quantum channel
initialization and key reconciliation, while providing stronger security
guarantees. PQC-only postures reduce setup cost but rely exclusively on
computational hardness assumptions. By separating handshake and per-message
overheads, the model clarifies how security cost scales with task graph structure
and communication intensity, supporting informed policy selection in large-scale,
distributed agentic AI systems.

\subsection{Security Trade-offs and Formalization in Quantum-Secure MAS}

In globally distributed Agentic AI systems, secure communication is anchored in three 
quantum-resilient technologies: Post-Quantum Cryptography (PQC), 
Quantum Key Distribution (QKD), and 
Quantum Random Number Generation (QRNG). 
PQC algorithms like Kyber and Dilithium offer high-performance encryption and digital signatures on standard hardware-A performance-optimized implementation by WolfSSL demonstrates that Kyber512 achieves an encapsulation time of approximately $5.25$ milliseconds on an ARM Cortex-M4 processor, yielding a throughput of $\approx 190$ encapsulations per second \cite{Ref013_Benchmark_wolfssl2024kyber} and, additionally, benchmarking results on a Raspberry Pi 3 confirm that Kyber512 significantly outperforms legacy cryptosystems, achieving encapsulation speeds hundreds of times faster than traditional ECDH implementations—highlighting its suitability for real-time, resource-constrained agentic deployments \cite{Ref014_Benchmark_10912602}. This makes PQC ideal for real-time agent messaging, key rotation, and distributed 
task execution over global IP networks. QKD, offering information-theoretic security, leverages quantum principles like no-cloning and entanglement. Twin-Field Quantum Key Distribution (QKD) demonstrates secure key rates of approximately $111.7$ kbps over fiber links spanning $202$ km, while state-of-the-art implementations have achieved up to $115.8$ Mbps over $10$ km and maintained secure transmissions across distances as long as $1,002$ km. Urban-scale QKD networks consistently operate with error rates below $0.65$\%, generating secret keys at rates exceeding $2 \times 10^{-5}$ per photon pulse \cite{Ref015_Benchmark_PhysRevLett.130.210801}\cite{Ref016_Benchmark_green2023qkd}\cite{Ref017_Benchmark_huang2025fullyconnected}. QRNG, complementing both, provides truly unpredictable entropy derived from quantum phenomena such as photon time-of-arrival or vacuum fluctuations. Commercial quantum random number generators (QRNGs), such as ID Quantique’s Quantis \cite{Ref12-idq}, produce up to $16$ Mbps of certified quantum entropy. More recent advancements, like Toshiba’s CMOS-integrated QRNG-on-chip module, deliver $8$ Mbps, enabling secure randomness generation in mobile, embedded, or edge-deployed agentic systems \cite{Ref018_Benchmark_idquantiqueQuantisQRNG}\cite{Ref019_Benchmark_marangon2024qrng}. 

Together, this triad forms a layered stack: QKD for ultra-sensitive communication, PQC for routine scalable coordination, and QRNG for cryptographic freshness, ensuring mission integrity, entropy fidelity, and future-proof resilience across edge, cloud, and interplanetary agent deployments. Table~\ref{tab:Table6} (in Appendix B) presents a comparison of security dimensions for Classical, PQC, and QKD+QRNG systems.

%% Section 7 Research Challenges and Future Vision
%%%%%%%%%%%%%%%%%%%%%%%%%%%%%%%%%%%%%%%%%%%%%%%%%%
\section{Research Challenges and Future Vision}

Several open challenges define the next frontier of secure, large-scale Agentic AI, including scalable quantum key distribution, quantum-native protocol stacks, cross-jurisdiction cryptographic policy integration, cryptography-aware AI reasoning, and agent self-sovereignty. Addressing these challenges is both a technical and strategic necessity for realizing agentic systems that are not only autonomous and intelligent, but also cryptographically trustworthy, compliant, and resilient in the post-quantum era.

\subsection{Scalable QKD: beyond point-to-point links}

Most existing quantum key distribution (QKD) deployments rely on point-to-point optical links between fixed nodes, which limits scalability in large, dynamic agentic systems distributed across heterogeneous networks. Enabling QKD at scale requires quantum networking architectures that support multi-hop key distribution through mechanisms such as quantum repeaters, entanglement swapping, and key relay
protocols. In this setting, agents must establish QKD-derived session keys without direct physical connectivity, relying on routed quantum resources and policy-governed trusted nodes. Achieving this further necessitates orchestration layers capable of managing quantum link availability, key bandwidth, session allocation, and security policy enforcement, which are essential for integrating QKD into globally distributed agentic AI ecosystems.

\subsection{Quantum IP stack: protocol layers for agent-level cryptographic APIs}

Integrating quantum security into agentic AI systems requires a quantum-native protocol stack that plays a role analogous to TCP/IP, but is designed for quantum-resilient communication and key management. This stack exposes layered
abstractions that secure agent interactions, including prompts, tool invocation, and memory access, while deriving ephemeral session keys via QKD or QRNG and establishing resilient transport channels using post-quantum cryptography. Agents interface with the stack through composable cryptographic APIs that dynamically select PQC, QKD, or hybrid modes based on runtime context and policy, enabling crypto-agnostic application logic while supporting auditability, key provenance, and graceful fallback in large-scale, heterogeneous deployments.

\subsection{Inter-jurisdictional crypto policy: cross-border use of quantum cryptography}

As agentic AI systems operate across national boundaries, they must comply with diverse cryptographic regulations, export controls, and data localization requirements. This necessitates policy-aware cryptographic negotiation, enabling
agents to select jurisdiction-compliant PQC schemes, activate regional QKD links, or rely on localized entropy sources when cloud-based QRNG is restricted. Support for key localization, legal metadata tagging, and jurisdiction-aware routing is therefore essential, requiring legal-policy constraints to be embedded directly into cryptographic decision-making and compliance auditing mechanisms.

\subsection{AI crypto-awareness: context-driven cryptographic decision-making}

In dynamic multi-agent environments, cryptography must transition from a passive service to an adaptive system capability. Crypto-aware agents dynamically select and adjust cryptographic postures based on mission context, threat assessment, and peer trust, escalating from PQC with quantum-derived entropy to QKD-secured communication when risk or sensitivity increases. Such behavior requires cryptographic state machines, real-time risk evaluation, and policy-aware control, embedding security decisions directly into agent planning, communication, and tool execution under evolving operational conditions.

\subsection{Agent self-sovereignty: decentralized identity and credential management}

In distributed agentic AI environments, such as autonomous fleets or decentralized financial systems, agents cannot depend on centralized identity authorities. Self-sovereign identity enables agents to manage their own cryptographic keys and credentials using decentralized identifiers and verifiable credentials bound to post-quantum key pairs and recorded on tamper-resistant ledgers. Trust and role verification are established through signed credentials exchanged peer-to-peer, supporting secure collaboration, fine-grained access control, and revocation without centralized oversight, while enhancing agent portability and long-term resilience.

%% Section 8 Experimentation
%%%%%%%%%%%%%%%%%%%%%%%%%%%%

\section{Empirical Validation}

The goal of the experimental evaluation is to demonstrate the feasibility and operational characteristics of the proposed architecture under present realistic deployment conditions. The experiments therefore focus on three aspects: (i) performance overhead introduced by quantum-secure cryptographic layers, where percent overhead appears large in micro-benchmarks, it reflects very small classical baselines; the absolute added latency remains in the low-millisecond range and is dominated by network latency in cloud-scale deployment.(ii) system robustness under adversarial conditions, and (iii) the practical integration of post-quantum and quantum-derived entropy sources within a distributed agentic infrastructure. The viability and performance of the proposed framework, centered on the layered cryptographic architecture shown in Figure~\ref{fig:Figure1}, were empirically validated through a two-phase evaluation combining controlled cryptographic micro-benchmarking and cloud-scale deployment analysis. 

A local research testbed was used to isolate the latency of post-quantum cryptographic primitives and quantum entropy injection, while a production-oriented Azure deployment was used to assess end-to-end behavior under realistic distributed conditions. In the cloud-scale measurements, PQC is executed natively, while QKD link latency is modeled from Phase I parameters because physical quantum fiber infrastructure is not currently available in standard cloud availability zones. Consistent with the cloud latency results, end-to-end transaction time is dominated by cloud networking ($\approx 297$ ms internal baseline), largely masking cryptographic overhead while preserving linear scaling as the worker pool increases. Across these experiments, the results show that the framework remains operationally feasible, preserves secure execution across the full trust boundary loop, and scales without altering the overall execution profile, while maintaining quantum-secure protections based on QRNG hardware \cite{IDQ2025}, standardized symmetric encryption \cite{ref_AES_NIST2001AES}, LangGraph-style orchestration \cite{ref_LangGraphDocs,ref_LangChain_LangChainIntro}, MCP-based coordination \cite{Ref012-MCP-hou2025mcp}, and modeled QKD-assisted links based on BB84 secure key rates \cite{Pirandola2020}.

%%%Table C.6

\textbf{\\Phase I: Cryptographic micro-benchmarking}\\

This phase focuses on isolating and quantifying the cryptographic latency introduced by the quantum-safe primitives.

\begin{table}[h]
\renewcommand{\thetable}{2} % set the display number
\centering
\caption{Comparative Performance and Size of PQC vs. Classical Cryptography. Metrics are grouped by key exchange performance, digital signature timing, and cryptographic artifact sizes.}
\label{tab:comparative_performance}
\scriptsize
\setlength{\tabcolsep}{4pt}
\begin{tabular}{lcccc}
\toprule
\textbf{Metric} & \textbf{PQC Mean} & \textbf{Classical Mean} & \textbf{Unit} & \textbf{Overhead (\%)} \\
\midrule
\hline
Encapsulation/Exchange & 0.082 & 0.018 & ms & 78.2 \\
Decapsulation & 0.014 & N/A & ms & N/A \\
\midrule
Key Generation & 0.11 & 0.01 & ms & 710.0 \\
Message Signing & 0.34 & 0.01 & ms & 2385.1 \\
Signature Verification & 0.08 & 0.03 & ms & 149.6 \\
\midrule
KEM Public Key Size & 1184 & 65 & bytes & 1721.5 \\
KEM Ciphertext Size & 1088 & N/A & bytes & N/A \\
Signature Public Key Size & 1952 & 32 & bytes & 6000.0 \\
Signature Size & 3309 & 64 & bytes & 5070.3 \\
\bottomrule
\end{tabular}
\end{table}

Analysis of raw primitives in Table~\ref{tab:comparative_performance} reveals the expected trade-off: PQC introduces a significant but bounded overhead compared to classical cryptography. The PQC implementation uses NIST-standardized key encapsulation and signature schemes, while the QKD layer models BB84 latency using a pragmatic short-haul secure key rate of 1 Mbps \cite{Pirandola2020}. The key hardware parameters for the QRNG and QKD layers are detailed in Table~\ref{tab:simulated_params}.

\begin{table}[h]
\renewcommand{\thetable}{3} % set the display number
\centering
\caption{Measured Hardware Latencies and Derivation Parameters}
\label{tab:simulated_params}
\scriptsize
\setlength{\tabcolsep}{4pt}
\begin{tabular}{llcc}
\toprule
\textbf{Layer} & \textbf{Parameter} & \textbf{Value} & \textbf{Unit}\\
\midrule
\hline
\multicolumn{4}{l}{\textbf{QRNG (Layer 2) Parameters - Measured}} \\
\midrule
& Nonce Generation Time & 1.157 & ms \\
& \textit{Observed Throughput} & $\approx 28$ & Mbps \\
\midrule
\multicolumn{4}{l}{\textbf{QKD (Layer 3, BB84 Protocol) Parameters - Simulated}} \\
\midrule
& Encrypt/Decrypt Time (AES) & 0.005 & ms \\
& Key Establishment Time & 10.0 & ms \\
& \textit{Key Size} & 128 & bits \\
& \textit{Implied Rate}  & 1 & Mbps \\
\bottomrule
\end{tabular}
\end{table}

The local testbed results for the seven critical channels are presented in Table~\ref{tab:7_channels_final_compressed} and visually summarized in Figure~\ref{fig:protocols_cost}.

\begin{table}[h]
\renewcommand{\thetable}{4} % set the display number
\centering
\caption{Performance Benchmark of Seven Critical Quantum-Secured Communication Channels}
\label{tab:7_channels_final_compressed}
\scriptsize
\setlength{\tabcolsep}{2.5pt}
\begin{tabular}{cl ccc rr}
\toprule
\textbf{ID} & \textbf{Communication Flow} & \textbf{PQC} & \textbf{QRNG} & \textbf{QKD} & \textbf{Time} & \textbf{Overhead}\\
 & & & & & \textbf{(ms)} & \textbf{(\%)}\\
\midrule
\hline
1 & User $\rightarrow$ Orch. (Request) & Y & Y & N & 1.06 & 6.6 \\
2 & Orch. $\leftrightarrow$ Agent (Handshake) & Y & N & N & 0.41 & 693.5 \\
3 & Orch. $\rightarrow$ MCP (Task Publish) & Y & Y & Y & 2.26 & 12.7 \\
4 & Agent $\leftarrow$ MCP (Fetch Task) & Y & N & Y & 1.10 & 8.8 \\
5 & Agent $\rightarrow$ MCP (Result) & Y & Y & Y & 3.38 & 68.8 \\
6 & Agent $\leftrightarrow$ Agent (Memory) & Y & Y & Y & 4.53 & 50.9 \\
7 & Orch. $\leftarrow$ MCP (Aggregation) & Y & N & Y & \textbf{9.83} & \textbf{16722.9} \\
\bottomrule
\multicolumn{7}{p{0.95\columnwidth}}{\scriptsize\textit{Note: PQC, QRNG, QKD application: Y=Yes, N=No.}}
\end{tabular}
\end{table}
.
%\begin{figure}[htbp]
\begin{figure}[H]
\setcounter{figure}{0}
\renewcommand{\thefigure}{2}
    \centering
    \includegraphics[width=0.70\textwidth]{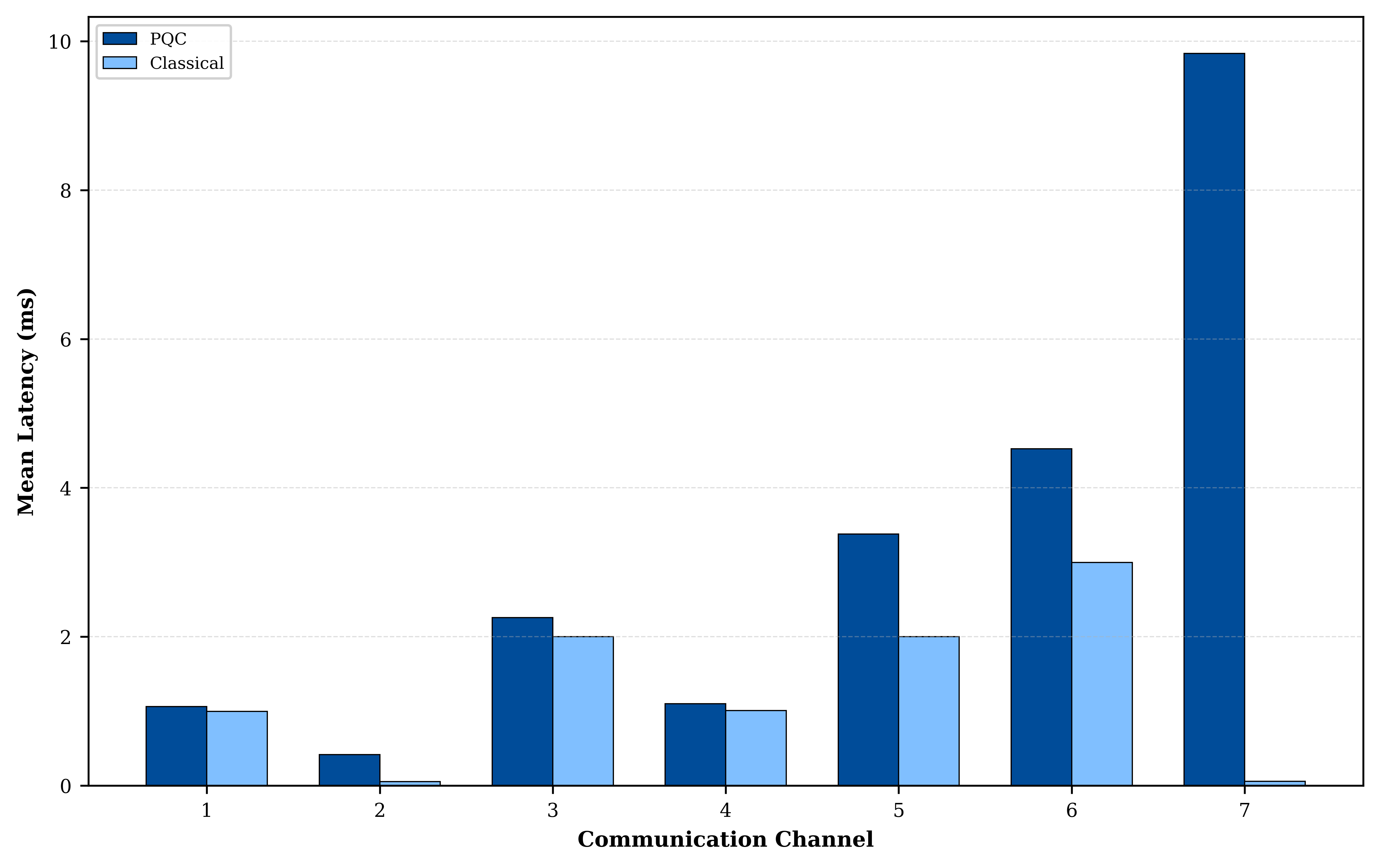}
    \caption{Absolute latency comparison of quantum-secured vs. classical protocols (ms).}
    \label{fig:protocols_cost}
\end{figure}

\textbf{\\Phase II: Cloud-scale integration and scalability} \\

To validate the framework under realistic network conditions, the full seven-channel architecture was benchmarked within the Azure environment. While PQC integrations were natively executed, QKD link latencies were mathematically modeled based on Phase I micro-benchmarks, since physical quantum fiber infrastructure is not currently available in standard cloud availability zones.

\begin{table}[h]
\renewcommand{\thetable}{5} % set the display number
\centering
\caption{Cloud-Scale Performance: Azure Deployment Latency (PQC-TLS 1.3 with Modeled QKD)}
\label{tab:cloud_latency}
\small
\setlength{\tabcolsep}{4pt}
\begin{tabular}{cl c c}
\toprule
\textbf{ID} & \textbf{Communication Channel} & \textbf{Security Stack} & \textbf{Mean Cloud Latency (ms)} \\
\midrule
\hline
1 & User $\rightarrow$ Orch. (Request) & PQC-TLS & 326.00 \\
2 & Orch. $\leftrightarrow$ Agent (Handshake) & PQC-TLS & 297.41 \\
3 & Orch. $\rightarrow$ MCP (Task Publish) & PQC-TLS + QKD & 299.26 \\
4 & Agent $\leftarrow$ MCP (Fetch Task) & PQC-TLS + QKD & 298.10 \\
5 & Agent $\rightarrow$ MCP (Result) & PQC-TLS + QKD & 300.38 \\
6 & Agent $\leftrightarrow$ Agent (Memory) & PQC-TLS + QKD & 301.53 \\
7 & Orch. $\leftarrow$ MCP (Aggregation) & PQC-TLS + QKD & 306.83 \\
\bottomrule
\end{tabular}
\end{table}

The cloud results in Table~\ref{tab:cloud_latency} show a consistent internal latency baseline of approximately 297 ms, indicating that cloud networking dominates total transaction time and largely masks the cryptographic overhead. As shown in Figure~\ref{fig:scalability_plots}, the PQC-enabled system maintains the same linear scaling profile as the classical baseline as the worker pool increases from $N=1$ to $N=50$, confirming that the sidecar-based security architecture does not introduce a serialization bottleneck during parallel execution.

%\begin{figure}[htbp]
\begin{figure}[H]
\setcounter{figure}{0}
\renewcommand{\thefigure}{3}
    \centering
    \includegraphics[width=0.9\textwidth]{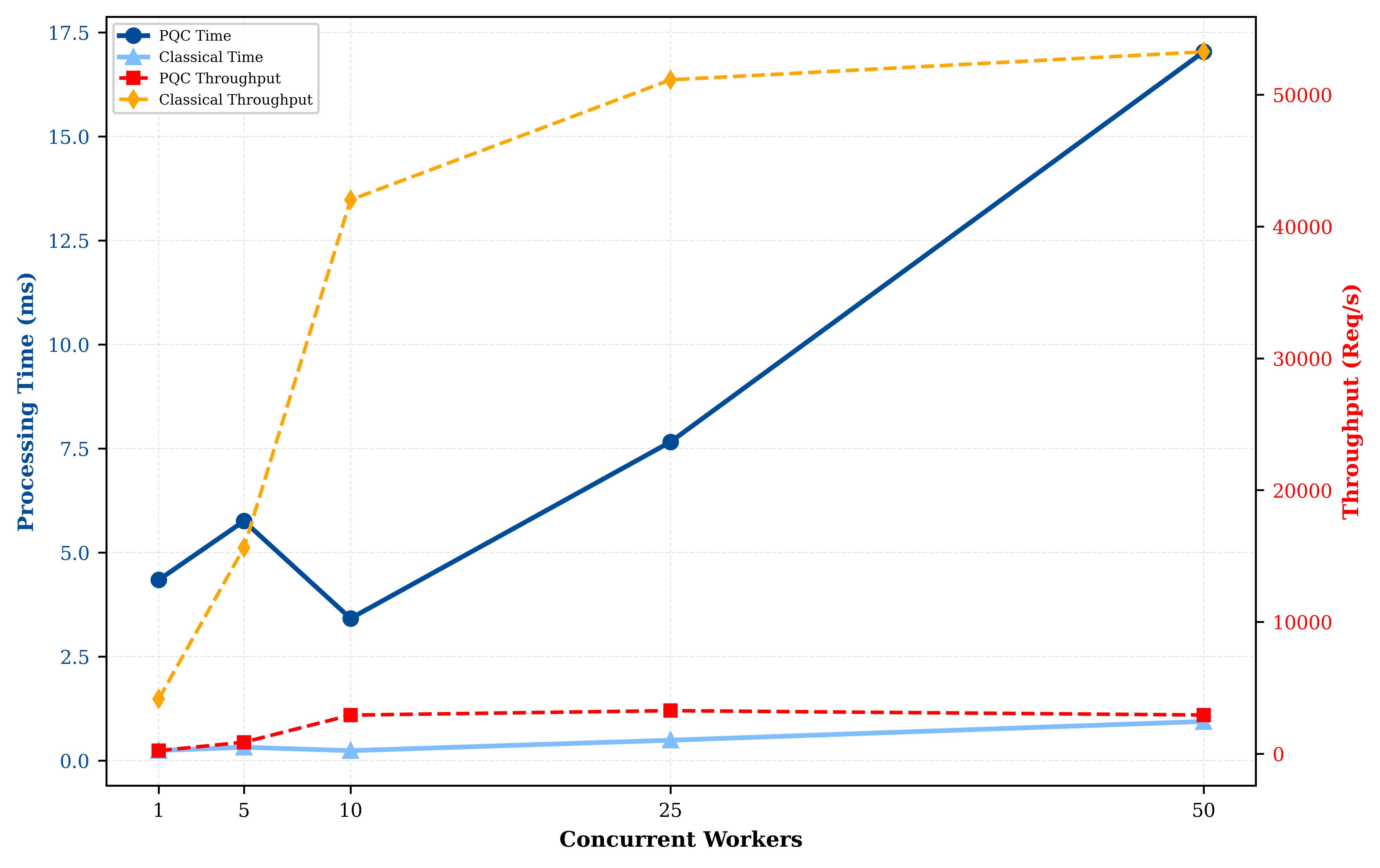}
    \caption{Scalability and throughput of the quantum-secured framework.}
    \label{fig:scalability_plots}
\end{figure}

The time for Channel 7 identifies the main cryptographic bottleneck in the current implementation. The Orchestrator sequentially verifies incoming agent signatures during aggregation, which was intentionally preserved in the testbed to isolate per-signature verification latency. In production, this bottleneck can be reduced through parallel verification or asynchronous batch processing. Figure~\ref{fig:Figure1} summarizes the layered architecture and trust-boundary structure used for empirical validation. Across the two-phase evaluation, the results indicate that the framework remains operationally feasible, preserves secure execution across distributed trust boundaries, and supports scalable multi-agent operation under the proposed quantum-secure-by-construction design. In a Monte Carlo adversarial simulation with 100,000 injected malicious vectors (tampering and replay), the framework achieved a 100 $\%$ detection/blocking rate via PQC signature verification and QRNG nonce-enforced freshness (Table \ref{tab:security_resilience_final} in Appendix C). Detailed experimental architecture, deployment protocol, hardware specifications, benchmark tables, cloud latency measurements, and adversarial simulation results are provided in Supplementary Information (Appendix C — Empirical validation details).

%% Section 9 Security Properties and Guarantees
%%%%%%%%%%%%%%%%%%%%%%%%%%%%%%%%%%%%%%%%%%%%%%%

\section{Security Properties and Guarantees}
\label{sec:security_properties}

This section summarizes the security guarantees provided by the proposed Quantum-Secured Agentic AI framework under standard cryptographic assumptions. Formal proofs are outside the scope of this work; instead, we state explicit security properties that follow from the construction and the underlying primitives.

\subsection{Confidentiality}

For links secured under post-quantum cryptographic postures, message confidentiality is achieved using an IND-CCA (Indistinguishability under Adaptive Chosen Ciphertext Attack)–secure post-quantum key encapsulation mechanism combined with authenticated encryption (AEAD). Under these assumptions, the advantage of any probabilistic polynomial-time adversary $\mathcal{A}$ in distinguishing encrypted messages is negligible:
\begin{equation}
\Pr[\mathcal{A} \text{ wins IND-CCA}] \leq \varepsilon_{\mathrm{pqc}}(\kappa),
\label{eq:confidentiality_indcca}
\end{equation}
where $\kappa$ denotes the security parameter, and $\varepsilon_{\mathrm{pqc}}$ is determined by the underlying PQC protocol parameters.

\subsection{Authenticity and integrity}

Message authenticity, integrity, and non-repudiation are ensured through post-quantum digital signature schemes secure against existential forgery under chosen-message attacks (EUF-CMA). Formally, for any adversary $\mathcal{A}$, the probability of producing a valid forgery is negligible:
\begin{equation}
\Pr[\mathcal{A} \text{ forges a valid signature}] \leq
\varepsilon_{\mathrm{sig}}(\kappa),
\label{eq:authenticity_eufcma}
\end{equation}
where $\kappa$ denotes the security parameter, and $\varepsilon_{\mathrm{sig}}$ is determined by the underlying signature protocol parameters.

These guarantees apply uniformly to inter-agent communication, task graph commitments, and audit log records, providing end-to-end accountability across the system.

\subsection{Information-theoretic security with QKD}

When quantum key distribution (QKD) is available, the framework attains information-theoretic security for session key material. Let $K$ denote a QKD-derived key and let $E$ represent all adversarial side information permitted by the threat model. Security guarantees ensure that the mutual information between $K$ and $E$ remains negligible:
\begin{equation}
I(K; E) \leq \varepsilon_{\mathrm{qkd}},
\label{eq:qkd_information_theoretic}
\end{equation}
where $\varepsilon_{\mathrm{qkd}}$ is determined by the underlying QKD protocol parameters.

\subsection{Compositional security}

By integrating post-quantum cryptographic primitives with QKD-derived entropy within a unified session key derivation mechanism, the framework achieves compositional security. Under this design, the compromise of any single cryptographic assumption does not trivially break system confidentiality or authenticity, provided that at least one underlying primitive remains secure. Together, these properties establish strong, future-resilient security guarantees for autonomous, distributed agentic AI systems operating across classical and quantum-enabled environments.

%% Section 10 Conclusion
%%%%%%%%%%%%%%%%%%%%%%%%

\section{Conclusion}

As Agentic AI systems expand across cloud, edge, robotic, and orbital platforms, secure communication becomes mission-critical. These globally distributed agents continuously exchange state, coordinate plans, and act autonomously over adversarial networks and fragmented legal regimes. Traditional cryptographic methods like RSA and ECC no longer suffice, vulnerable to both quantum threats and modern attacks such as impersonation or data poisoning. To address this, we propose a modular, quantum-secure architecture tailored for global Agentic AI. It integrates three layers: Post-Quantum Cryptography (PQC) as a high-performance baseline, Quantum Random Number Generators (QRNG) for non-deterministic entropy, and Quantum Key Distribution (QKD) for provable, forward-secure communication. These technologies work together to secure the entire agent lifecycle, from identity and orchestration to memory and tools. This adaptable framework enables PQC for low-latency coordination, QKD for high-assurance paths, and QRNG-seeded keys where entropy auditability is critical. Rather than treating cryptography as a backend service, it places cryptographic resilience at the core of agent design,  supporting verifiable identity, synchronized state, and trusted coordination across diverse environments, from terrestrial data centers to high-latency space links. Quantum-secure architectures are no longer aspirational but they are foundational. This blueprint ensures that Agentic AI systems can remain autonomous, intelligent, and trustworthy even in the face of emerging quantum and geopolitical threats. Beyond proposing a conceptual framework, this work demonstrates that integrating quantum-safe security mechanisms directly into the architecture of agentic AI systems is technically feasible and operationally practical within modern cloud-based infrastructures.

%% References
%%%%%%%%%%%%%

\bibliographystyle{unsrt}  
\bibliography{Paper}

%% Appendix
%%%%%%%%%%%
\newpage
%\appendix{\textbf{Appendix A — Quantum Secured Multi Agent System Algorithms} \\}
\textbf{Appendix A — Quantum Secured Multi Agent System Algorithms}
%%%%%%%%%%%%%%%% Algorithm 1: QuantumSecuredMultiAgentSystem() %%%%%%%%%%%%
%%%%%%%%%%%%%%%%%%%%%%%%%%%%%%%%%%%%%%%%%%%%%%%%%%%%%%%%%%%%%%%%%%%%%%%%%%%

%\hline
\noindent\rule{\textwidth}{0.4pt}
\textbf{ \\ Algorithm 1: QuantumSecuredMultiAgentSystem()} \\
\noindent\rule{\textwidth}{0.4pt}
%\hline
\textbf{Purpose: \\} \\
To securely orchestrate task execution in a geo-distributed, multi-agent system spanning cloud, edge, and physical systems using layered quantum cryptographic protocols (PQC, QRNG, and QKD). \\

\textbf{Input:}

Let:

\begin{itemize}
    \item $\mathcal{A} = \{A_{1}, A_{2}, \dots, A_{n}\}$: Set of participating specialist agents (regional, physical, cloud, edge)
    \item $O$: Central Orchestrator
    \item $C$: Client (device/browser/agent)
    \item $Q_{\mathrm{QRNG}}$: Quantum Random Number Generator
    \item $Q_{\mathrm{QKD}}$: Quantum Key Distribution overlay (optional)
    \item $K_{\mathrm{PQC}}$: Post-Quantum Cryptography stack:
    \begin{itemize}
        \item Kyber (KEM) for key exchange
        \item Dilithium for digital signatures
        \item AES-GCM for symmetric encryption
    \end{itemize}
\end{itemize}

\textbf{Output:}

\begin{itemize}
    \item $\mathcal{R}$: Secure, verifiable system response to client task
    \item $\mathcal{L}$: Immutable audit trail with PQC signatures, QRNG tokens, QKD status
\end{itemize}

\textbf{Definitions:}

\begin{itemize}
    \item $\mathcal{S}$: Secure session context initialized via \textsc{InitializeSession}()
    \item $TG$: Task DAG created by orchestrator in \textsc{BuildTaskGraph}()
    \item $L$: Log entries signed and persisted via \textsc{RecordAuditTrail}()
    \item $A_{i}.\textsc{Execute}()$: Agent-local task execution step in \textsc{SecureAgentExecution}()
    \item $M$: Memory (federated, append-only, geo-replicated)
\end{itemize}

%%%%%%%%%% algorithm steps %%%%%%%%%%%%%
\textbf{Algorithm Steps:}

%%\BlankLine
\textbf{Step 0: Bootstrap secure session} \\
$\mathcal{S} \leftarrow$ InitializeSession($C, O, \mathcal{A}, Q_{\mathrm{QRNG}}, Q_{\mathrm{QKD}}, K_{\mathrm{PQC}}$) \\

\textbf{Step 1: Receive and secure user request} \\
$\mathrm{Req}_{C} \leftarrow$ SecureClientRequest({$C, O, \mathcal{S}, Q_{\mathrm{QRNG}}, Q_{\mathrm{QKD}}, K_{\mathrm{PQC}}$}) \\

\textbf{Step 2: Build the secure task graph with least privilege} \\
$TG \leftarrow$ BuildTaskGraph({$O, \mathrm{Req}_{C}, M, Q_{\mathrm{QRNG}}, K_{\mathrm{PQC}}, Q_{\mathrm{QKD}}$}) \\

\textbf{Step 3: Execute graph securely across agents} \\
For Each{subtask $\tau \in TG$}{
    $AssignedAgent \leftarrow \tau.\mathrm{assignee}$ \\
    $Result_{\tau} \leftarrow$ SecureAgentExecution({$AssignedAgent, \tau, Q_{\mathrm{QRNG}}, K_{\mathrm{PQC}}, Q_{\mathrm{QKD}}$}) \\
    $M.\mathrm{write}(Result_{\tau}, encrypted=\mathrm{True}, signed=\mathrm{True})$
} \\

\textbf{Step 4: Merge and reason over agent results} \\
$FinalOutput \leftarrow$ MergeAndReason({$O, M, Q_{\mathrm{QRNG}}, K_{\mathrm{PQC}}, Q_{\mathrm{QKD}}$}) \\

\textbf{Step 5: Final policy sweep and response} \\
$\mathcal{R} \leftarrow$ FinalizeAndRespond({$O, FinalOutput, C, Q_{\mathrm{QRNG}}, Q_{\mathrm{QKD}}, K_{\mathrm{PQC}}$}) \\

\textbf{Step 6: Record full execution trace immutably} \\
$\mathcal{L} \leftarrow$ RecordAuditTrail({$O, \mathcal{A}, Q_{\mathrm{QRNG}}, Q_{\mathrm{QKD}}, K_{\mathrm{PQC}}$}) \\

\textbf{Step 7: Return} \\
return({$(\mathcal{R}, \mathcal{L})$}) \\

%%\end{algorithm}

%%%%%%%%%%%%%%%% Algorithm 2: Quantum Secured Session Bootstrap() %%%%%%%%%%%%
%%%%%%%%%%%%%%%%%%%%%%%%%%%%%%%%%%%%%%%%%%%%%%%%%%%%%%%%%%%%%%%%%%%%%%%%%%%%%%%%%%
\noindent\rule{\textwidth}{0.4pt}
%\hline
\textbf{ \\ Algorithm 2: Quantum Secured Session Bootstrap()}\\
\noindent\rule{\textwidth}{0.4pt}
%\hline
\textbf{Purpose: \\} \\
To establish a secure communication session between distributed actors (agents, orchestrator, clients) using Post-Quantum Cryptography (PQC), Quantum Random Number Generators (QRNG), and Quantum Key Distribution (QKD) when available.

\textbf{Input:}
\begin{itemize}
    \item $\mathcal{A} = \{A_{1}, A_{2}, \dots, A_{n}\}$: Set of agents
    \item $O$: Orchestrator
    \item $C$: Client (edge or cloud)
    \item $Q_{\mathrm{QRNG}}$: Quantum random number source
    \item $Q_{\mathrm{QKD}}$: Quantum Key Distribution infrastructure (optional)
    \item $K_{\mathrm{PQC}}$: PQC key management system supporting:
    \begin{itemize}
        \item Kyber (KEM) for key establishment
        \item Dilithium for digital signature verification
        \item AES-GCM for symmetric encryption
    \end{itemize}
\end{itemize}

\textbf{Output:}
$\mathcal{S} = (\textit{session\_key}, \textit{nonce}, \textit{qkd\_id}, \textit{sig\_certs})$ securely initialized between the parties

\textbf{Definitions:}
\begin{itemize}
    \item $\mathrm{TEE}(A_{i})$: Trusted Execution Environment proof for agent $A_{i}$
    \item $\mathrm{Cert}_{\mathrm{Dilithium}}(A_{i})$: PQC signature certificate of agent $A_{i}$
    \item $\textit{nonce} \sim Q_{\mathrm{QRNG}}$: Nonce derived from quantum entropy
    \item $QKD_{i,j}$: QKD-derived symmetric key between nodes $i$ and $j$
\end{itemize}

\textbf{Algorithm Steps: \\} \\

\textbf{Step 1: Iterate} \\
For each actor {$X \in \{C, O\} \cup \mathcal{A}$} \\

\hspace*{1em} 
\textbf{Step 1.1: Identity and Signature Setup} \\
    \hspace*{1em} 
    Load PQC identity certificate: $\mathrm{Cert}_{\mathrm{Dilithium}}(X)$ \; \\
    \hspace*{1em}
    Verify $\mathrm{Cert}_{\mathrm{Dilithium}}(X)$ using the PQ-safe CA chain \; \\
    \hspace*{1em}
    If $X$ is a physical/edge agent \\
    \hspace*{2em} 
        verify $\mathrm{TEE}(X)$ \; \\
    \hspace*{1em}
    End If \\

\hspace*{1em}
\textbf{Step 1.2: Quantum Entropy Bootstrapping} \\
    \hspace*{1em}
    $nonce_{X} \leftarrow Q_{\mathrm{QRNG}}.\mathrm{sample}()$ \; \\
    \hspace*{1em}
    $session\_id_{X} \leftarrow Q_{\mathrm{QRNG}}.\mathrm{sample}()$ \; \\
    \hspace*{1em}
    $device\_key\_seed \leftarrow Q_{\mathrm{QRNG}}.\mathrm{sample}()$ \; \\
    \hspace*{1em}
    If{$Q_{\mathrm{QRNG}}$ unavailable}{ \\
    \hspace*{2em} 
        use DRBG fallback and log ``degraded entropy'' \;  \\
    \hspace*{1em}
    End If \\
    
\hspace*{1em}
\textbf{Step 1.3: Key Establishment: QKD or PQC } \\
    \hspace*{1em}
    If QKD link available between $X$ and orchestrator \\
    \hspace*{2em} 
        $key_{X} \leftarrow \mathrm{QKD\_derive}(X, O)$ \; \\
    \hspace*{2em} 
        $qkd\_session\_id_{XO} \leftarrow \mathrm{QKD.session\_id}(X, O)$ \; \\
    %}
    \hspace*{1em}
    Else \\
    \hspace*{2em} 
        tcp{PQC fallback: Kyber KEM handshake} \\
    \hspace*{2em} 
        $key_{X} \leftarrow \mathrm{Kyber\_Encapsulate}(pubkey_{O})$ \; \\
    \hspace*{2em} 
        $secret_{X} \leftarrow \mathrm{Kyber\_Decapsulate}(privkey_{X}, key_{X})$ \; \\
    \hspace*{2em} 
        $qkd\_session\_id_{XO} \leftarrow \mathrm{null}$ \; \\
    \hspace*{1em}
    End If \\

\hspace*{1em}
\textbf{Step 1.4: Publish Capability Map for Coordination} \\
    \hspace*{1em}
    $caps_{X} \leftarrow \{\; \\
    \hspace*{2em} 
        pqc\_support: \mathrm{true},\; \\
    \hspace*{2em} 
        qkd\_support: \mathrm{QKD.available}(X),\; \\
    \hspace*{2em} 
        sig\_algo: \text{"DILITHIUM3"},\; \\
    \hspace*{2em} 
        enc\_algo: \text{"KYBER768+AES256-GCM"}\; \}$ \; \\
    
\hspace*{1em}
\textbf{Step 1.5: Emit Signed Bootstrapping Event} \\
    \hspace*{1em}
    $\mathrm{AuditLog.append}(\{ \\
    \hspace*{2em} 
        event: \text{"BOOTSTRAP\_INIT"},\; \\
    \hspace*{2em} 
        actor: X,\; \\
    \hspace*{2em} 
        timestamp: \mathrm{now}(),\; \\
    \hspace*{2em} 
        cert: \mathrm{Cert}_{\mathrm{Dilithium}}(X),\; \\
    \hspace*{2em} 
        qkd\_status: caps_{X}.qkd\_support,\; \\
    \hspace*{2em} 
        qrng\_nonce: nonce_{X},\; \\
    \hspace*{2em} 
        session\_id: session\_id_{X} \}, \\ \;
    \hspace*{2em} 
        signed\_by=\mathrm{Cert}_{\mathrm{Dilithium}}(X))$ \; \\
end For \\

\textbf{Step 2: Return} \\
Return {$\mathcal{S} := \{
\hspace*{1em} 
    session\_keys: \{key_{X} \ \forall X\},\;
\hspace*{1em} 
    qkd\_map: \{qkd\_session\_id_{XO}\},\;
\hspace*{1em} 
    nonces: \{nonce_{X}\},\;
\hspace*{1em} 
    certs: \{\mathrm{Cert}_{\mathrm{Dilithium}}(X)\}
\}$} \\
%%\end{algorithm}

%%%%%%%%%%%%%%%% Algorithm 3: InitializeSession() %%%%%%%%%%%%%%
%%%%%%%%%%%%%%%%%%%%%%%%%%%%%%%%%%%%%%%%%%%%%%%%%%%%%%%%%%%%%%%%%%%%

%\hline
\noindent\rule{\textwidth}{0.4pt}
\textbf{ \\ Algorithm 3: InitializeSession()} \\
\noindent\rule{\textwidth}{0.4pt}
%\hline
\textbf{Purpose: \\} \\
Securely initialize the communication session between the orchestrator, clients, and agents using quantum cryptographic primitives: PQC (Kyber, Dilithium), QRNG, and optionally QKD.

\textbf{Input:}
\begin{itemize}
    \item $\mathcal{A} = \{A_{1}, A_{2}, \dots, A_{n}\}$: Set of participating agents (cloud, regional, physical)
    \item $O$: Orchestrator
    \item $C$: Client (Edge app / service / system)
    \item $Q_{\mathrm{QRNG}}$: Quantum Random Number Generator service
    \item $Q_{\mathrm{QKD}}$: QKD infrastructure (optional; for trusted backbone)
    \item $K_{\mathrm{PQC}}$: Post-quantum key management system supporting:
    \begin{itemize}
        \item Kyber (KEM): for key exchange
        \item Dilithium: for digital signatures
        \item AES-GCM: for symmetric encryption
    \end{itemize}
\end{itemize}

\textbf{Output:}
\begin{itemize}
    \item $\mathcal{S} = (\textit{session\_keys}, \textit{nonces}, \textit{qkd\_links}, \textit{certs})$: Session context for all actors
    \item Signed bootstrapping audit events in the append-only audit ledger
\end{itemize}

\textbf{Definitions:}
\begin{itemize}
    \item $\mathrm{TEE}(X)$: Trusted Execution Environment attestation for agent $A_{i}$
    \item $\mathrm{Cert}_{\mathrm{Dilithium}}(X)$: PQ-safe signature certificate of actor $X$
    \item $\mathrm{QRNG.sample}()$: Entropy-derived sample for nonces/session IDs
    \item $QKD_{i,j}$: Symmetric key derived from QKD between $i$ and $j$
\end{itemize}

\textbf{Algorithm Steps:} \\ \\
\textbf{Step 1: Iterate \\}
For each actor $X \in \{C, O\} \cup \mathcal{A}$} \\ \\
    \hspace*{1em}
    \textbf{Step 1.1: Identity Verification} \\
    \hspace*{1em}
    $cert_{X} \leftarrow \mathrm{Cert}_{\mathrm{Dilithium}}(X)$\; \\
    \hspace*{1em}
    Verify $cert_{X}$ against PQC CA chain\; \\
    \hspace*{1em}
    If {$X$ is edge/physical} \\
        \hspace*{2em}
        assert $\mathrm{TEE}(X) == \mathrm{valid}$\; \\
    \hspace*{1em}
    End If \\ \\
    \hspace*{1em}
    \textbf{Step 1.2: Quantum Randomness Bootstrapping} \\
    \hspace*{1em}
    $nonce_{X} \leftarrow Q_{\mathrm{QRNG}}.\mathrm{sample}()$\; \\
    \hspace*{1em}
    $session\_id_{X} \leftarrow Q_{\mathrm{QRNG}}.\mathrm{sample}()$\; \\
    \hspace*{1em}
    $device\_key_{X} \leftarrow Q_{\mathrm{QRNG}}.\mathrm{sample}()$\; \\
    \hspace*{1em}
    If{$Q_{\mathrm{QRNG}}$ fails} \\
    \hspace*{1em}
        $device\_key_{X} \leftarrow \mathrm{DRBG}.\mathrm{sample}()$\; \\
        \hspace*{2em}
        log ``degraded entropy'' for $X$\; \\ \\
    \hspace*{1em}
    \textbf{Step 1.3: Secure Channel Establishment} \\
    \hspace*{1em}
    If QKD link exists between $X$ and $O$ \\
        \hspace*{2em}
        $key_{X} \leftarrow \mathrm{QKD.derive}(X, O)$\; \\
        \hspace*{2em}
        $qkd\_id_{X} \leftarrow \mathrm{QKD.session\_id}(X, O)$\; \\
    \hspace*{1em}
    Else \\ 
        \hspace*{2em}
        \textbf{Step: 1.3.1: Fallback to PQC Kyber} \\
        \hspace*{2em}
        $pk_{O} \leftarrow \mathrm{GetPublicKey}(O)$\; \\
        \hspace*{2em}
        $key_{X} \leftarrow \mathrm{Kyber\_Encapsulate}(pk_{O})$\; \\
        \hspace*{2em}
        $shared\_secret_{X} \leftarrow \mathrm{Kyber\_Decapsulate}(privkey_{X}, key_{X})$\; \\
        \hspace*{2em}
        $qkd\_id_{X} \leftarrow \mathrm{null}$\; \\
    \hspace*{1em}
    End If \\ \\
    \hspace*{1em}
    \textbf{Step 1.4: Capability Declaration} \\
    \hspace*{1em}
    $caps_{X} \leftarrow \{ \\
        \hspace*{2em}
        pqc\_support: \mathrm{true},\; \\
        \hspace*{2em}
        qkd\_support: \mathrm{QKD.available}(X),\; \\
        \hspace*{2em}
        sig\_algo: \text{"DILITHIUM3"},\; \\
        \hspace*{2em}
        enc\_algo: \text{"KYBER768+AES256-GCM"}\}$\; \\ \\
    \hspace*{1em}
    \textbf{Step 1.5: Signed Audit Log Event} \\
    \hspace*{1em}
    $\mathrm{AuditLog.append}(\{ \\
        \hspace*{2em}
        event: \text{"BOOTSTRAP\_INIT"},\; \\
        \hspace*{2em}
        actor: X,\; \\
        \hspace*{2em}
        cert: cert_{X},\; \\
        \hspace*{2em}
        timestamp: \mathrm{now}(),\; \\
        \hspace*{2em}   
        qrng\_nonce: nonce_{X},\; \\
        \hspace*{2em}
        session\_id: session\_id_{X},\; \\
        \hspace*{2em}
        qkd\_status: caps_{X}.qkd\_support \\
        \hspace*{2em}
    \},\; signed\_by = cert_{X})$\; \\ \\
end For

\textbf{Step 2: Return} \\
Return {$\mathcal{S} = \{
    \hspace*{1em}
    session\_keys: \{key_{X}\},\; \\
    \hspace*{1em}
    nonces: \{nonce_{X}\},\; \\
    \hspace*{1em}
    qkd\_links: \{qkd\_id_{X}\},\; \\
    \hspace*{1em}
    certs: \{cert_{X}\}\}$} \\

%%%%%%%%%%%%%%%% Algorithm 4: SecureClientRequest() %%%%%%%%%%%%%%%%
%%%%%%%%%%%%%%%%%%%%%%%%%%%%%%%%%%%%%%%%%%%%%%%%%%%%%%%%%%%%%%%%%%%%
%\hline
\noindent\rule{\textwidth}{0.4pt}
\textbf{\\ Algorithm 4: SecureClientRequest()} \\
\noindent\rule{\textwidth}{0.4pt}
%\hline
%%%%%%%%%%%%%%%%%%% Algorithm Purpose %%%%%%%%%%%%%%%%%%%%%
\textbf{Purpose: \\} \\
Establish a quantum-secure communication between the client and the orchestrator to initiate a multi-agent task request. Ensures data confidentiality, origin authenticity, and tamper-evident communication using PQC, QRNG, and optionally QKD.

%%%%%%%%%%%%%%%%%%% Algorithm Inputs %%%%%%%%%%%%%%%%%%%%%
\textbf{Input:}
\begin{itemize}
    \item $\mathcal{G}(V, E)$: Task graph from Algorithm~3
    \item $\rho: V \rightarrow \mathcal{A}$: Mapping from tasks to assigned agents
    \item $Q_{\mathrm{QRNG}}$: Quantum entropy source for per-task \textit{request\_id} and \textit{write\_token}
    \item $K_{\mathrm{PQC}}$: Post-Quantum Cryptography stack:
    \begin{itemize}
        \item Kyber for key encapsulation
        \item Dilithium for digital signatures
        \item AES-GCM for local encryption
    \end{itemize}
    \item $Q_{\mathrm{QKD}}$: Optional QKD sessions between orchestrator and agents
    \item $\mathrm{SharedMemory}$: Append-only context store
    \item $\mathrm{AgentCerts}$: PQ-safe public key certificates for all agents (Dilithium)
    \item $\mathrm{session\_context}$: Includes PQC keys, QKD channels, and QRNG-generated session identifiers
\end{itemize}

%%%%%%%%%%%%%%%%%%% Algorithm Outputs %%%%%%%%%%%%%%%%%%%%%
\textbf{Output:}
\begin{itemize}
    \item $\mathrm{SecureArtifact}$: PQC/QKD-protected agent result written to shared memory with full lineage metadata
    \item $\mathrm{AuditRecord}$: Agent invocation trace with QRNG linkage and Dilithium signature
\end{itemize}

%%%%%%%%%%%%%%%%%%% Algorithm Definitiona %%%%%%%%%%%%%%%%%%%%%
\textbf{Definitions:}
\begin{itemize}
    \item $\mathrm{Execute}(v_{s}, A_{i})$: Executes subtask $v_{s}$ on agent $A_{i}$
    \item $\mathrm{QKD.key}(i, j)$: Symmetric key between node $i$ and $j$ if QKD session exists
    \item $\mathrm{QRNG.sample}()$: Fresh entropy sample for ID, token, or nonce
    \item $\mathrm{Encrypt}(\textit{data}, \textit{key})$: PQC or QKD symmetric encryption
    \item $\mathrm{Sign}(\textit{data}, \textit{privkey})$: Generate Dilithium digital signature
    \item $\mathrm{TEE}(A_{i})$: Trusted Execution Environment attestation (for physical agents)
    \item $\mathrm{Log.append}(\textit{event})$: Signed audit logging operation
\end{itemize}

%%%%%%%%%%%%%%%%%%% Algorithm Steps %%%%%%%%%%%%%%%%%%%%%
\textbf{Algorithm Steps:} \\
\textbf{\\ Step 1: Iterate} \\
For Each {$v_{s} \in V$} \\
    \hspace*{1em}
    $A_{i} \leftarrow \rho(v_{s})$ tcp{Assigned agent} \\
    \hspace*{1em}
    $cert_{A_{i}} \leftarrow \mathrm{AgentCerts}[A_{i}]$ \; \\ \\
    \hspace*{1em}
    \textbf{Step 1.1: Transport Security Setup} \\
    \hspace*{1em}
    If {$\mathrm{QKD.key}(O, A_{i})$ available} \\
        \hspace*{2em}
        $session\_key \leftarrow \mathrm{QKD.key}(O, A_{i})$ \; \\
        \hspace*{2em}
        $qkd\_session\_id \leftarrow \mathrm{QKD.session\_id}(O, A_{i})$ \; \\
    \hspace*{1em}
    Else \\
        \hspace*{2em}
        $session\_key \leftarrow \mathrm{Kyber\_Encapsulate}(pubkey_{A_{i}})$ \; \\
        \hspace*{2em}
        $qkd\_session\_id \leftarrow \mathrm{null}$ \; \\
    \hspace*{1em}
    End If \\
    \hspace*{1em}
    $request\_id \leftarrow Q_{\mathrm{QRNG}}.\mathrm{sample}()$ \; \\
    \hspace*{1em}
    $nonce \leftarrow Q_{\mathrm{QRNG}}.\mathrm{sample}()$ \; \\ \\
    \hspace*{1em}
    \textbf{Step 1.2: Execute Task} \\
    \hspace*{1em}
    $result \leftarrow \mathrm{Execute}(task = v_{s}, agent = A_{i})$ \; \\ \\
    \hspace*{1em}
    \textbf{Step 1.3: Provenance \& Encryption} \\
    \hspace*{1em}
    $provenance \leftarrow \{ \\
        \hspace*{2em}
        task\_id: v_{s}.task\_id,\; \\
        \hspace*{2em}
        request\_id: request\_id,\; \\
        \hspace*{2em}
        agent: A_{i},\; \\
        \hspace*{2em}
        timestamp: \mathrm{now}(),\; \\
        \hspace*{2em}
        qkd\_session\_id: qkd\_session\_id,\; \\
        \hspace*{2em}
        nonce: nonce \}$ \; \\
    \hspace*{1em}
    $encrypted\_result \leftarrow \mathrm{Encrypt}(result, session\_key)$ \; \\
    \hspace*{1em}
    $signature \leftarrow \mathrm{Sign}(encrypted\_result \ \| \ provenance, priv_{A_{i}})$ \; \\ \\
    \hspace*{1em}
    \textbf{Step 1.4: Persist to Shared Memory} \\
    \hspace*{1em}
    $write\_token \leftarrow Q_{\mathrm{QRNG}}.\mathrm{sample}()$ \; \\
    \hspace*{1em}
    $\mathrm{SharedMemory.append}(\{ \\
        \hspace*{2em}
        payload: encrypted\_result,\; \\
        \hspace*{2em}
        signature: signature,\; \\
        \hspace*{2em}
        metadata: provenance,\; \\
        \hspace*{2em}
        write\_token: write\_token \})$ \; \\ \\
    \hspace*{1em}
    \textbf{Step 1.5: Emit Audit Log} \\
    \hspace*{1em}
    $\mathrm{AuditLog.append}(\{ \\
        \hspace*{2em}
        event: \text{"AGENT\_EXECUTION\_COMPLETED"},\; \\
        \hspace*{2em}
        agent: A_{i},\; \\
        \hspace*{2em}
        task\_id: v_{s}.task\_id,\; \\
        \hspace*{2em}
        request\_id: request\_id,\; \\
        \hspace*{2em}
        write\_token: write\_token,\; \\
        \hspace*{2em}
        timestamp: \mathrm{now}(),\; \\
        \hspace*{2em}
        sig: signature \})$ \; \\
End For \\
%%%%%%%%%%%%%%%% Algorithm: 5 BuildTaskGraph() %%%%%%%%%%%%%%%%
%%%%%%%%%%%%%%%%%%%%%%%%%%%%%%%%%%%%%%%%%%%%%%%%%%%%%%%%%%%%%%%
%\hline
\noindent\rule{\textwidth}{0.4pt}
\textbf{\\ Algorithm 5: BuildTaskGraph()} \\
\noindent\rule{\textwidth}{0.4pt}
%\hline
%%%%%%%%%%%%%%%%%%% Algorithm Purpose %%%%%%%%%%%%%%%%%%%%%
\textbf{Purpose: \\} \\
Decompose the user request into a parallelizable task graph, assign agents by capability and region, and generate quantum-secure identifiers for full traceability. This serves as the core orchestration step in a geo-distributed multi-agent system.

%%%%%%%%%%%%%%%%%%% Algorithm Inputs %%%%%%%%%%%%%%%%%%%%%
\begin{itemize}
    \item $\mathrm{secure\_request}$: The encrypted, signed, and nonce-tagged user request from Algorithm~2
    \item $O$: Orchestrator agent
    \item $\mathcal{A} = \{A_{1}, A_{2}, \dots, A_{n}\}$: Registered specialist agents across cloud, regional, and physical locations
    \item $Q_{\mathrm{QRNG}}$: Quantum random number generator
    \item $K_{\mathrm{PQC}}$: PQ-safe crypto stack (Kyber, Dilithium, AES-GCM)
    \item $\mathrm{SharedMemory}$: Contextual store for user/system state and access policies
    \item $\mathrm{session\_context}$: Bootstrap metadata from Algorithm~0: session keys, nonces, agent certificates, QKD sessions
    \item $\mathrm{CapabilityMap}$: Each agent’s declared capabilities, location, and cryptographic availability
\end{itemize}

%%%%%%%%%%%%%%%%%%% Algorithm Outputs %%%%%%%%%%%%%%%%%%%%%
\textbf{Output:}
\begin{itemize}
    \item A task graph $\mathcal{G}(V, E)$: Directed acyclic graph of subtasks with quantum-secure IDs
    \item Task-to-agent assignment map $\rho: V \rightarrow \mathcal{A}$
    \item Dilithium-signed audit event (\texttt{"TASK\_GRAPH\_CREATED"})
\end{itemize}

%%%%%%%%%%%%%%%%%%% Algorithm Definitiona %%%%%%%%%%%%%%%%%%%%%
\textbf{Definitions:}
\begin{itemize}
    \item $\mathrm{Verify}(\textit{sig}, \textit{pub}_{C})$: Validates Dilithium signature using client certificate
    \item $\mathrm{Decrypt}(\textit{payload}, \textit{session\_key})$: Decrypts request using PQC or QKD session key
    \item $\mathrm{QRNG.sample}()$: Quantum entropy source for ID/nonce generation
    \item $\mathcal{G}(V, E)$: Task graph with vertices $V$ and edges $E$
    \item $\mathrm{AgentCaps}(A_{i})$: Declared capabilities of agent $A_{i}$
    \item $\mathrm{SharedMemory.query}()$: Federated context lookup
    \item $\mathrm{AuditLog.append}()$: Tamper-proof, append-only log operation
\end{itemize}

%%%%%%%%%%%%%%%%%%% Algorithm Steps %%%%%%%%%%%%%%%%%%%%%
\textbf{Algorithm Steps:} \\

\textbf{Step 1: Intake request and decrypt} \\
Validate Dilithium signature from $\mathrm{secure\_request}$\; \\
$session\_key \leftarrow session\_context.session\_keys[C]$\; \\
$task\_payload \leftarrow \mathrm{Decrypt}(\mathrm{secure\_request.payload}, session\_key)$\; \\
Validate QRNG nonce freshness $\leftarrow$ Check against replay DB\; \\

\textbf{Step 2: Extract user/system context} \\
$user\_context \leftarrow \mathrm{SharedMemory.query}(user\_id = C)$\; \\
$policy\_constraints \leftarrow user\_context.access\_policy$\; \\ \\

\textbf{Step 3: Intent detection and task parsing} \\
$intent \leftarrow \mathrm{Parse}(task\_payload)$\; \\
$subtasks \leftarrow \mathrm{Decompose}(intent)$ tcp*{NLP/Planning model} \\
$constraints \leftarrow \mathrm{ExtractConstraints}(task\_payload, user\_context)$\; \\

\textbf{Step 4: Construct task graph} \\
Initialize graph $\mathcal{G}$ with vertices $V = \emptyset$, edges $E = \emptyset$\; \\
For Each {$s \in subtasks$} \\
    \hspace*{1em}
    $task\_id_{s} \leftarrow Q_{\mathrm{QRNG}}.\mathrm{sample}()$\; \\
    \hspace*{1em}
    $A \leftarrow \mathrm{SelectAgent}(s, CapabilityMap, policy\_constraints)$\; \\
    \hspace*{1em}
    Add vertex $v_{s} = (task\_id_{s}, s, A)$ to $V$\; \\
End For \\
Add edges $E$ based on execution dependencies (DAG rules)\; \\

\textbf{Step 5: Finalize routing metadata} \\
For Each {$v \in V$} \\
    \hspace*{1em}
    $v.metadata.nonce \leftarrow Q_{\mathrm{QRNG}}.\mathrm{sample}()$\; \\
    \hspace*{1em}
    $v.metadata.qkd\_session \leftarrow \mathrm{QKD.session\_if\_available}(A_{i}, O)$\; \\
    \hspace*{1em}
    $v.metadata.policy \leftarrow policy\_constraints$\; \\
End For \\

\textbf {Step 6: Record provenance} \\
$graph\_hash \leftarrow \mathrm{Hash}(\mathcal{G})$\; \\
$\mathrm{AuditLog.append}(\{ \\
    \hspace*{1em}
    event: \text{"TASK\_GRAPH\_CREATED"},\; \\
    \hspace*{1em}
    orchestrator: O,\; \\
    \hspace*{1em}
    task\_ids: [v.task\_id \ \mathrm{for} \ v \in V],\; \\
    \hspace*{1em}
    constraints: constraints,\; \\
    \hspace*{1em}
    provenance: graph\_hash,\; \\
    \hspace*{1em}
    timestamp: \mathrm{now}(),\; \\
    \hspace*{1em}
    sig: \mathrm{Dilithium\_Sign}(graph\_hash, priv_{O}) \})$\; \\

\textbf{Step 7: Return} \\
Return{$(\mathcal{G}(V, E), \rho)$} \\

%%%%%%%%%%%%%%%% Algorithm 6: SecureAgentExecution() %%%%%%%%%%%%%%%%
%%%%%%%%%%%%%%%%%%%%%%%%%%%%%%%%%%%%%%%%%%%%%%%%%%%%%%%%%%%%%%%%%%%%%
%\hline
\noindent\rule{\textwidth}{0.4pt}
\textbf{\\ Algorithm 6: SecureAgentExecution()} \\
\noindent\rule{\textwidth}{0.4pt}
%\hline
%%%%%%%%%%%%%%%%%%% Algorithm Purpose %%%%%%%%%%%%%%%%%%%%%
\textbf{Purpose: \\} \\
Execute a given subtask on a selected agent (regional, physical, or cloud-based) with quantum-secure communication and provenance. The agent must interact with its local systems, call APIs/tools, and store results securely while maintaining tamper-resistance and compliance.

%%%%%%%%%%%%%%%%%%% Algorithm Inputs %%%%%%%%%%%%%%%%%%%%%
\textbf{Input:}
\begin{itemize}
    \item $\mathcal{G}(V, E)$: Task graph from Algorithm~5
    \item $\rho: V \rightarrow \mathcal{A}$: Mapping from tasks to assigned agents
    \item $Q_{\mathrm{QRNG}}$: Quantum entropy source for per-task \textit{request\_id} and \textit{write\_token}
    \item $K_{\mathrm{PQC}}$: Post-Quantum Cryptography stack:
    \begin{itemize}
        \item Kyber for key encapsulation
        \item Dilithium for digital signatures
        \item AES-GCM for local encryption
    \end{itemize}
    \item $Q_{\mathrm{QKD}}$: Optional QKD sessions between orchestrator and agents
    \item $\mathrm{SharedMemory}$: Append-only context store
    \item $\mathrm{AgentCerts}$: PQ-safe public key certificates for all agents (Dilithium)
    \item $\mathrm{session\_context}$: Includes PQC keys, QKD channels, and QRNG-generated session identifiers
\end{itemize}

%%%%%%%%%%%%%%%%%%% Algorithm Outputs %%%%%%%%%%%%%%%%%%%%%
\textbf{Output:}
\begin{itemize}
    \item $\mathrm{SecureArtifact}$: PQC/QKD-protected agent result written to shared memory with full lineage metadata
    \item $\mathrm{AuditRecord}$: Agent invocation trace with QRNG linkage and Dilithium signature
\end{itemize}

%%%%%%%%%%%%%%%%%%% Algorithm Definitiona %%%%%%%%%%%%%%%%%%%%%
\textbf{Definitions:}
\begin{itemize}
    \item $\mathrm{Execute}(v_{s}, A_{i})$: Executes subtask $v_{s}$ on agent $A_{i}$
    \item $\mathrm{QKD.key}(i, j)$: Symmetric key between node $i$ and $j$ if QKD session exists
    \item $\mathrm{QRNG.sample}()$: Fresh entropy sample for ID, token, or nonce
    \item $\mathrm{Encrypt}(\textit{data}, \textit{key})$: PQC or QKD symmetric encryption
    \item $\mathrm{Sign}(\textit{data}, \textit{privkey})$: Generate Dilithium digital signature
    \item $\mathrm{TEE}(A_{i})$: Trusted Execution Environment attestation (for physical agents)
    \item $\mathrm{Log.append}(\textit{event})$: Signed audit logging operation
\end{itemize}

%%%%%%%%%%%%%%%%%%% Algorithm Steps %%%%%%%%%%%%%%%%%%%%%
\textbf{Algorithm Steps:} \\

\textbf{Step 1: Iterate} \\
For Each {$v_{s} \in V$} \\
    \hspace*{1em}
    $A_{i} \leftarrow \rho(v_{s})$ tcp {Assigned agent} \\
    \hspace*{1em}
    $cert_{A_{i}} \leftarrow \mathrm{AgentCerts}[A_{i}]$ \; \\

    \hspace*{1em}
    \textbf{Step 1.1: Transport Security Setup} \\
    \hspace*{1em}
    If{$\mathrm{QKD.key}(O, A_{i})$ available} \\
        \hspace*{2em}
        $session\_key \leftarrow \mathrm{QKD.key}(O, A_{i})$ \; \\
        \hspace*{2em}
        $qkd\_session\_id \leftarrow \mathrm{QKD.session\_id}(O, A_{i})$ \; \\
    \hspace*{1em}
    Else \\
        \hspace*{2em}
        $session\_key \leftarrow \mathrm{Kyber\_Encapsulate}(pubkey_{A_{i}})$ \; \\
        \hspace*{2em}
        $qkd\_session\_id \leftarrow \mathrm{null}$ \; \\
    \hspace*{1em}
    End If \\
    \hspace*{1em}
    $request\_id \leftarrow Q_{\mathrm{QRNG}}.\mathrm{sample}()$ \; \\
    \hspace*{1em}
    $nonce \leftarrow Q_{\mathrm{QRNG}}.\mathrm{sample}()$ \; \\
    
    \hspace*{1em}
    \textbf{Step 1.2: Execute Task} \\
    \hspace*{1em}
    $result \leftarrow \mathrm{Execute}(task = v_{s}, agent = A_{i})$ \; \\
    
    \hspace*{1em}
    \textbf{Step 1.3: Provenance \& Encryption} \\
    \hspace*{1em}
    $provenance \leftarrow \{ \\
        \hspace*{2em}
        task\_id: v_{s}.task\_id,\; \\
        \hspace*{2em}
        request\_id: request\_id,\; \\
        \hspace*{2em}
        agent: A_{i},\; \\
        \hspace*{2em}
        timestamp: \mathrm{now}(),\; \\
        \hspace*{2em}
        qkd\_session\_id: qkd\_session\_id,\; \\
        \hspace*{2em}
        nonce: nonce \}$ \; \\
    \hspace*{1em}
    $encrypted\_result \leftarrow \mathrm{Encrypt}(result, session\_key)$ \; \\
    \hspace*{1em}
    $signature \leftarrow \mathrm{Sign}(encrypted\_result \ \| \ provenance, priv_{A_{i}})$ \; \\

    \hspace*{1em}
    \textbf{Step 1.4: Persist to Shared Memory} \\
    \hspace*{1em}
    $write\_token \leftarrow Q_{\mathrm{QRNG}}.\mathrm{sample}()$ \; \\
    \hspace*{1em}
    $\mathrm{SharedMemory.append}(\{ \\
        \hspace*{2em}
        payload: encrypted\_result,\; \\
        \hspace*{2em}
        signature: signature,\; \\
        \hspace*{2em}
        metadata: provenance,\; \\
        \hspace*{2em}
        write\_token: write\_token \})$ \; \\

    \hspace*{1em}
    \textbf{Step 1.5: Emit Audit Log} \\
    \hspace*{1em}
    $\mathrm{AuditLog.append}(\{ \\
        \hspace*{2em}
        event: \text{"AGENT\_EXECUTION\_COMPLETED"},\; \\
        \hspace*{2em}
        agent: A_{i},\; \\
        \hspace*{2em}
        task\_id: v_{s}.task\_id,\; \\
        \hspace*{2em}
        request\_id: request\_id,\; \\
        \hspace*{2em}
        write\_token: write\_token,\; \\
        \hspace*{2em}
        timestamp: \mathrm{now}(),\; \\
        \hspace*{2em}
        sig: signature \})$ \; \\
End For \\

%%%%%%%%%%%%%%%% Algorithm 7: MergeAndReason() %%%%%%%%%%%%%%%%
%%%%%%%%%%%%%%%%%%%%%%%%%%%%%%%%%%%%%%%%%%%%%%%%%%%%%%%%%%%%%%%
%\hline
\noindent\rule{\textwidth}{0.4pt}
\textbf{\\ Algorithm 7: MergeAndReason()} \\
\noindent\rule{\textwidth}{0.4pt}
%\hline
%%%%%%%%%%%%%%%%%%% Algorithm Purpose %%%%%%%%%%%%%%%%%%%%%
\textbf{Purpose: \\} \\
Consolidate results from distributed agents, validate signatures and QRNG-based provenance, apply rules/logic/LLM planning, and produce structured, traceable outputs with cryptographic assurances.

%%%%%%%%%%%%%%%%%%% Algorithm Inputs %%%%%%%%%%%%%%%%%%%%%
\textbf{Input:}
\begin{itemize}
    \item $\mathcal{A} = \{a_{1}, a_{2}, \dots, a_{n}\}$: Set of all participating agents
    \item $\mathrm{SharedMemory}$: Append-only global context store with PQC-encrypted, signed artifacts
    \item $Q_{\mathrm{QRNG}}$: Quantum entropy source
    \item $K_{\mathrm{PQC}}$: PQ-safe cryptographic stack:
    \begin{itemize}
        \item Dilithium (signature verification)
        \item Kyber + AES-GCM (envelope decryption)
    \end{itemize}
    \item $\mathrm{Ruleset}\ R$: Logical constraints and orchestration policy rules (SLO, PII, data jurisdiction, budget)
    \item $\mathrm{Planner}$: Reasoning engine (symbolic, statistical, or LLM-based)
    \item $\mathrm{QKD}$: Optional symmetric keys from QKD sessions
    \item $\mathrm{session\_context}$: Signature certificates, expected QRNG lineage, and key metadata
\end{itemize}

%%%%%%%%%%%%%%%%%%% Algorithm Outputs %%%%%%%%%%%%%%%%%%%%%
\textbf{Output:}
\begin{itemize}
    \item $\mathrm{FinalReport}$: Structurally validated, signed, and encrypted output
    \item $\mathrm{AuditEvent}$: Cryptographically signed log of reasoning phase with lineage metadata
\end{itemize}

%%%%%%%%%%%%%%%%%%% Algorithm Definitiona %%%%%%%%%%%%%%%%%%%%%
\textbf{Definitions:}
\begin{itemize}
    \item $\mathrm{Decrypt}(\textit{payload}, \textit{key})$: Use Kyber-wrapped AES-GCM or QKD key to decrypt result
    \item $\mathrm{VerifySig}(\textit{data}, \textit{sig}, \textit{pubkey})$: Validate Dilithium3 signature of a message
    \item $\mathrm{QRNG.verify\_linkage}(\textit{task\_id}, \textit{nonce})$: Confirm entropy lineage for replay prevention
    \item $\mathrm{Hash}(\textit{data})$: Cryptographic hash (e.g., SHA3-256)
    \item $\mathrm{Provenance}(\textit{data})$: Source $\rightarrow$ transform $\rightarrow$ output graph for data item
    \item $\mathrm{Planner.infer}(\dots)$: Apply logic/LLM over secure inputs and constraints
\end{itemize}

%%%%%%%%%%%%%%%%%%% Algorithm Steps %%%%%%%%%%%%%%%%%%%%%
\textbf{Algorithm Steps:} \\ \\
Inputs $\leftarrow \emptyset$

\textbf{Step 1: Retrieve and verify results from memory}

For Each {$v_{s} \in$ completed tasks in $\mathcal{A}$} \\
    \hspace*{1em}
    $record \leftarrow \mathrm{SharedMemory.fetch}(task\_id = v_{s}.task\_id)$\; \\

    \hspace*{1em}
    \textbf{Step 1.1: Validate signature} \\
    \hspace*{1em}
    $is\_valid \leftarrow \mathrm{VerifySig}(record.payload \ \| \ record.metadata,\ record.signature,\ \mathrm{AgentCerts}[record.agent])$\; \\
    \hspace*{1em}
    If {$\lnot is\_valid$}
        \hspace*{2em}
        Raise ``SignatureMismatchError''\; \\
    \hspace*{1em}
    End If \\

    \hspace*{1em}
    \textbf{Step 1.2: Validate QRNG-based replay resistance} \\
    \hspace*{1em}
    If {$\lnot \mathrm{QRNG.verify\_linkage}(record.task\_id, record.metadata.nonce)$} \\
        \hspace*{2em}
        Raise ``EntropyReplayDetected''\; \\
    \hspace*{1em}
    End If \\ \\
    \hspace*{1em}
    \textbf{Step 1.3: Decrypt the payload} \\
    \hspace*{1em}
    If {$\mathrm{QKD.key}(record.agent, Orchestrator)$ available} \\
        \hspace*{2em}
        $key \leftarrow \mathrm{QKD.key}(record.agent, Orchestrator)$\; \\
    \hspace*{1em}
    Else \\
        \hspace*{2em}
        $key \leftarrow \mathrm{Kyber\_Decapsulate}(priv_{Orchestrator},\ record.payload.kem)$\; \\
    \hspace*{1em}
    End If \\
    \hspace*{1em}
    $plaintext \leftarrow \mathrm{Decrypt}(record.payload, key)$\; \\
    \hspace*{1em}
    $Inputs \leftarrow Inputs \cup \{ (v_{s}.task\_id,\ plaintext,\ \mathrm{provenance}(record)) \}$\; \\
End For \\

\textbf{Step 2: Constraint-aware reasoning} \\
$Rules \leftarrow \mathrm{Load}(policy\_ruleset\ R)$\; \\
$Output \leftarrow \mathrm{Planner.infer}(Inputs,\ constraints = Rules)$\; \\

\textbf{Step 3: Prepare final output} \\
$summary \leftarrow \{ \\
    \hspace*{1em}
    human\_summary: Output.human\_text,\; \\
    \hspace*{1em}
    machine\_output: Output.json,\; \\
    \hspace*{1em}
    provenance\_hash: \mathrm{Hash}(Inputs),\; \\
    \hspace*{1em}
    timestamp: \mathrm{now}() \}$\;

$final\_ciphertext \leftarrow \mathrm{Encrypt}(summary, session\_context.final\_key)$\; \\
$signature \leftarrow \mathrm{Sign}(summary, priv_{Orchestrator})$\; \\

\textbf{Step 4: Persist and audit} \\
$\mathrm{SharedMemory.append}(\{ \\
    \hspace*{1em}
    payload: final\_ciphertext,\; \\
    \hspace*{1em}
    signature: signature,\; \\
    \hspace*{1em}
    metadata: \{ \\
        \hspace*{2em}
        type: \text{"MERGE\_RESULT"},\; \\
        \hspace*{2em}
        task\_ids: [v.task\_id\ \mathrm{for}\ v \in Inputs],\; \\
        \hspace*{2em}
        timestamp: \mathrm{now}(),\; \\
        \hspace*{2em}
        nonce: Q_{\mathrm{QRNG}}.\mathrm{sample}() \} \})$\; \\

\textbf{Step 5: Audit Log} \\
$\mathrm{AuditLog.append}(\{ \\
    \hspace*{1em}
    event: \text{"MERGE\_REASONING\_COMPLETED"},\; \\
    \hspace*{1em}
    timestamp: \mathrm{now}(),\; \\
    \hspace*{1em}
    provenance\_hash: \mathrm{Hash}(Inputs),\; \\
    \hspace*{1em}
    task\_count: |Inputs|,\; \\
    \hspace*{1em}
    sig: signature \})$\; \\

\textbf{Step 6: Return} \\
Return {$Output$} \\

%%%%%%%%%%%%%%%% Algorithm 8: FinalizeAndRespond() %%%%%%%%%%%%
%%%%%%%%%%%%%%%%%%%%%%%%%%%%%%%%%%%%%%%%%%%%%%%%%%%%%%%%%%%%%%%%%%%

%\hline
\noindent\rule{\textwidth}{0.4pt}
\textbf{\\ Algorithm 8: FinalizeAndRespond()} \\
\noindent\rule{\textwidth}{0.4pt}
%\hline
%%%%%%%%%%%%%%%%%%% Algorithm Purpose %%%%%%%%%%%%%%%%%%%%%
\textbf{Purpose: \\} \\
Performs integrity verification, enforces policy compliance (e.g., residency, PII), and delivers cryptographically secure responses to the client. This is the final output sealing and delivery stage.

%%%%%%%%%%%%%%%%%%% Algorithm Inputs %%%%%%%%%%%%%%%%%%%%%
\textbf{Input:}
\begin{itemize}
    \item $\mathrm{Output}$: Final report from \textsc{MergeAndReason}() (includes human-readable and machine-parsable content)
    \item $\mathrm{Client}\ C$: The original requester (edge device, browser, system, etc.)
    \item $\mathrm{session\_context}$: Contains:
    \begin{itemize}
        \item $\mathrm{certs}$: PQC certificates
        \item $\mathrm{session\_keys}$: Per-agent session keys (Kyber/KDF or QKD)
        \item $\mathrm{nonces}$: QRNG nonces used in earlier phases
        \item $\mathrm{qkd\_map}$: Backbone QKD sessions
    \end{itemize}
    \item $\mathrm{Ruleset}\ R$: Data use, privacy, and residency policies
    \item $Q_{\mathrm{QRNG}}$: Quantum entropy source (for new nonce)
    \item $K_{\mathrm{PQC}}$: PQ-safe crypto stack (Kyber, Dilithium, AES-GCM)
\end{itemize}

%%%%%%%%%%%%%%%%%%% Algorithm Outputs %%%%%%%%%%%%%%%%%%%%%
\textbf{Output:}
\begin{itemize}
    \item $\mathrm{SecureResponse}$: A PQC-encrypted, signed, nonce-bound message sent to the client
    \item $\mathrm{FinalAuditEvent}$: A signed log verifying the delivery and integrity constraints
\end{itemize}

%%%%%%%%%%%%%%%%%%% Algorithm Definitiona %%%%%%%%%%%%%%%%%%%%%
\textbf{Definitions:}
\begin{itemize}
    \item $\mathrm{ValidateSignatures}(\textit{data})$: Check that all contributing agents signed their results
    \item $\mathrm{CheckPolicy}(\textit{tags})$: Enforce policies such as ``no cross-border PII transfer''
    \item $\mathrm{Encrypt}(\textit{data}, \textit{key})$: AES-GCM sealed with Kyber or QKD-derived key
    \item $\mathrm{Sign}(\textit{data}, \textit{priv})$: Digital signature with Dilithium
    \item $\mathrm{Nonce} \leftarrow Q.\mathrm{sample}()$: Fresh entropy from QRNG
\end{itemize}

%%%%%%%%%%%%%%%%%%% Algorithm Steps %%%%%%%%%%%%%%%%%%%%%
\textbf{Algorithm Steps:} \\ \\
\textbf{Step 1: Integrity sweep} \\
For Each {$artifact \in Output.inputs\_used$} \\
    \hspace*{1em}
    If {$\lnot \mathrm{ValidateSignatures}(artifact)$} \\
        \hspace*{2em}
        Raise ``InvalidSignatureException''\; \\
    \hspace*{1em}
    End If \\
    \hspace*{1em}
    If {$\mathrm{QRNG.verify\_linkage}(artifact.task\_id, artifact.nonce) == \mathrm{False}$} \\
        \hspace*{2em}
        Raise ``ReplayDetectedError''\; \\
    \hspace*{1em}
    End If \\
End For \\

\textbf{Step 2: Policy checks} \\
$metadata\_tags \leftarrow \mathrm{extract}(Output.policy\_tags)$\; \\
If {$\lnot \mathrm{CheckPolicy}(metadata\_tags, Ruleset\ R)$} \\
    \hspace*{1em}
    Raise ``PolicyViolationError''\; \\
End If \\

\textbf{Step 3: Prepare secure response} \\
$final\_nonce \leftarrow Q_{\mathrm{QRNG}}.\mathrm{sample}()$\; \\
If {$\mathrm{QKD.key}(Client, Orchestrator)$ available} \\
    \hspace*{1em}
    $response\_key \leftarrow \mathrm{QKD.key}(Client, Orchestrator)$\; \\
    \hspace*{1em}
    $qkd\_flag \leftarrow \text{"QKD\_ACTIVE"}$\; \\
Else \\
    \hspace*{1em}
    $response\_key \leftarrow \mathrm{Kyber\_Encapsulate}(pubkey_{C})$\; \\
    \hspace*{1em}
    $qkd\_flag \leftarrow \text{"PQC\_ONLY"}$\; \\
End If

$response\_payload \leftarrow \mathrm{Encrypt}(Output, response\_key)$\; \\
$response\_sig \leftarrow \mathrm{Sign}(response\_payload \ \| \ final\_nonce,  priv_{Orchestrator})$\;

$\mathrm{SecureResponse} \leftarrow \{ \\
    \hspace*{1em}
    task\_id: Output.task\_id,\; \\
    \hspace*{1em}
    payload: response\_payload,\; \\
    \hspace*{1em}
    nonce: final\_nonce,\; \\
    \hspace*{1em}
    signature: response\_sig,\; \\
    \hspace*{1em}
    qkd\_status: qkd\_flag,\; \\
    \hspace*{1em}
    timestamp: \mathrm{now}() \}$\; \\

\textbf{Step 4: Send and log} \\
$\mathrm{Transport.send}(Client.endpoint, \mathrm{SecureResponse})$\; \\
$\mathrm{AuditLog.append}(\{ \\
    \hspace*{1em}
    event: \text{"FINAL\_RESPONSE\_SENT"},\; \\
    \hspace*{1em}
    task\_id: Output.task\_id,\; \\
    \hspace*{1em}
    to: Client.id,\; \\
    \hspace*{1em}
    timestamp: \mathrm{now}(),\; \\
    \hspace*{1em}
    sig: response\_sig,\; \\
    \hspace*{1em}
    nonce: final\_nonce,\; \\
    \hspace*{1em}
    qkd\_used: qkd\_flag \})$\; \\

\textbf{Step 5: Post-response cleanup} \\
Schedule key rotation: $session\_key[C] \leftarrow \mathrm{rotate}()$\; \\
Clear any transient memory contexts\; \\

\textbf{Step 6: Return} \\
Return {$\mathrm{SecureResponse}$} \\

%%%%%%%%%%%%%%%% Algorithm 9: RecordAuditTrail()%%%%%%%%%%%%%%%%
%%%%%%%%%%%%%%%%%%%%%%%%%%%%%%%%%%%%%%%%%%%%%%%%%%%%%%%%%%%%%%%%%%%%

%\hline
\noindent\rule{\textwidth}{0.4pt}
\textbf{\\ Algorithm 9: RecordAuditTrail()} \\
\noindent\rule{\textwidth}{0.4pt}
%\hline
%%%%%%%%%%%%%%%%%%% Algorithm Purpose %%%%%%%%%%%%%%%%%%%%%
\textbf{Purpose: \\} \\
To ensure every cryptographically significant event within the multi-agent system is immutably logged, quantum-randomly linked, cryptographically signed, and—when available—replicated across sites via QKD-secured channels.

%%%%%%%%%%%%%%%%%%% Algorithm Inputs %%%%%%%%%%%%%%%%%%%%%
\textbf{Input:}
\begin{itemize}
    \item $E = \{e_{1}, e_{2}, \dots, e_{k}\}$: Event stream emitted from all actors (Orchestrator, Agents, Memory, etc.)
    \item $Q_{\mathrm{QRNG}}$: Quantum entropy source
    \item $\mathrm{QKD}$: QKD network overlay across sites (optional)
    \item $K_{\mathrm{PQC}}$: PQ-safe cryptography stack:
    \begin{itemize}
        \item Dilithium: Digital signature algorithm
        \item AES-GCM: Symmetric authenticated encryption
        \item Kyber: Key encapsulation mechanism (for object-level encryption)
    \end{itemize}
\end{itemize}

%%%%%%%%%%%%%%%%%%% Algorithm Outputs %%%%%%%%%%%%%%%%%%%%%
\textbf{Output:}
\begin{itemize}
    \item $\mathrm{AuditLog}$: Append-only ledger with quantum-secured, tamper-evident records replicated across geographically distributed sites
\end{itemize}

%%%%%%%%%%%%%%%%%%% Algorithm Definitiona %%%%%%%%%%%%%%%%%%%%%
\textbf{Definitions:}
\begin{itemize}
    \item $qrng\_nonce \leftarrow Q.\mathrm{sample}()$: Fresh quantum-derived randomness for event ID correlation
    \item $\mathrm{Sign}(\textit{event}, \textit{privkey})$: Digital signature using Dilithium
    \item $\mathrm{Hash}(\textit{payload})$: Cryptographic hash for content integrity (e.g., SHA3-256)
    \item $\mathrm{QKD.replication\_key}(\textit{src}, \textit{dst})$: Symmetric key from QKD channel (if exists)
\end{itemize}

%%%%%%%%%%%%%%%%%%% Algorithm Steps %%%%%%%%%%%%%%%%%%%%%
\textbf{Algorithm Steps:} \\

\textbf{Step 1: Iterate} \\
For Each {$e \in E$} \\ \\
    \hspace*{1em}
    \textbf{Step 1.1: Entropy binding and correlation} \\
    \hspace*{1em}
    $e.event\_id \leftarrow \text{"qrng:e-"} \ \| \ Q.\mathrm{sample}()$\; \\
    \hspace*{1em}
    $e.correlation\_id \leftarrow \mathrm{Hash}(e.task\_id \ \| \ e.event\_type \ \| \ Q.\mathrm{sample}())$\; \\ \\
    \hspace*{1em}
    \textbf{Step 1.2: Signing and hashing} \\
    \hspace*{1em}
    $e.payload\_hash \leftarrow \mathrm{Hash}(e.payload)$\; \\
    \hspace*{1em}
    $e.signature \leftarrow \mathrm{Sign}(e.payload \ \| \ e.event\_id \ \| \ e.timestamp,\ priv\_{actor}(e.source))$\; \\ \\
    \hspace*{1em}
    \textbf{Step 1.3: Append to local ledger} \\
    \hspace*{1em}
    $\mathrm{Ledger.append}(\{ \\
        \hspace*{2em}
        event\_id: e.event\_id,\; \\
        \hspace*{2em}
        source: e.source,\; \\
        \hspace*{2em}
        type: e.event\_type,\; \\
        \hspace*{2em}
        timestamp: e.timestamp,\; \\
        \hspace*{2em}
        sig: e.signature,\; \\
        \hspace*{2em}
        hash: e.payload\_hash,\; \\
        \hspace*{2em}
        correlation\_id: e.correlation\_id,\; \\
        \hspace*{2em}
        qkd\_status: \mathrm{QKD.available}(e.source, e.destination) \})$\; \\ \\
    \hspace*{1em}
    \textbf{Step 1.4: Replication (if required)} \\
    \hspace*{1em}
    If {$e.requires\_replication$} \\
        \hspace*{2em}
        For Each {$replica\_site \in e.replicas$} \\
            \hspace*{3em}
            If {$\mathrm{QKD.link}(e.source, replica\_site)$} \\
                \hspace*{4em}
                $key\_{rep} \leftarrow \mathrm{QKD.replication\_key}(e.source, replica\_site)$\; \\
                \hspace*{4em}
                $channel \leftarrow \mathrm{AES\!-\!GCM}(channel\_key = key\_{rep})$\; \\
                \hspace*{4em}
                $payload \leftarrow channel.encrypt(e)$\; \\
                \hspace*{4em}
                $flag \leftarrow \text{"QKD\_REPL"}$\; \\
            \hspace*{3em}
            Else \\
                \hspace*{4em}
                $key\_{rep} \leftarrow \mathrm{Kyber\_Encapsulate}(pubkey\_{replica})$\; \\
                \hspace*{4em}
                $payload \leftarrow \mathrm{AES\!-\!GCM}(key\_{rep}).encrypt(e)$\; \\
                \hspace*{4em}
                $flag \leftarrow \text{"PQC\_FALLBACK"}$\; \\
            \hspace*{3em}
            End If \\
            \hspace*{3em}
            $\mathrm{ReplicaLedger}[replica\_site].\mathrm{append}(\{ \\
                \hspace*{4em}
                source\_event: e.event\_id,\; \\
                \hspace*{4em}
                payload: payload,\; \\
                \hspace*{4em}
                qkd\_flag: flag,\; \\
                \hspace*{4em}
                ts: \mathrm{now}() \})$\; \\
        \hspace*{2em}
        End For \\
    \hspace*{1em}
    End If \\ \\
    \hspace*{1em}
    \textbf{Step 1.5: Flag degradations} \\
    \hspace*{1em}
    If {$Q.\mathrm{unavailable}()$} \\
        \hspace*{2em}
        $\mathrm{EmitAlert}(\text{"DEGRADED\_ENTROPY"},\ actor = e.source)$\; \\
    \hspace*{1em}
    End If \\
    \hspace*{1em}
    If {$\mathrm{QKD.link\_failure}(e.source, \mathrm{any}(replica\_site))$} \\
        \hspace*{2em}
        $\mathrm{EmitAlert}(\text{"DEGRADED\_QKD"},\ link = (e.source \rightarrow replica\_site))$\; \\
    \hspace*{1em}
    End If \\
End For \\ \\
\textbf{Step 2: Return} \\
Return {$\mathrm{Ledger}$} \\

%\appendix{\textbf{Appendix B — Domain Mapping of Quantum-Secured MAS Use Cases} \\}
\textbf{Appendix B — Domain Mapping of Quantum-Secured MAS Use Cases}\\

Table B.1 represents this final step enforces privacy-by-design and provides verifiable assurances for
high-assurance deployments, including financial, defense, and healthcare applications.

\begin{table}[H]
\renewcommand{\thetable}{B.1} % set the display number
    \centering
    \caption{Mapping of Algorithm Steps to Architecture Sections, Roles, and Quantum Security Controls}
    \begin{tabularx}{\textwidth}{XXXX}
    \hline
    \textbf{Operational Step} & \textbf{Architecture Section(s)} & \textbf{Agent Subsystem / Orchestrator Role} & \textbf{Quantum Security Controls} \\ 
    \hline
    \hline
    Quantum Secured Agentic AI System & All Layers (Global Coordination Layer, Agent Layer, Quantum Cryptographic Stack) &
    All orchestrators and agent subsystems across the pipeline &
    Full-stack PQC (Kyber, Dilithium), QRNG-sampled state transitions, QKD (if available) \\ 
    \hline

    Quantum Secured Session Bootstrap & Global Coordination Layer \newline Quantum Cryptographic Stack &
    Orchestrator: Session Manager, Crypto Negotiator &
    Session key exchange (Kyber), agent identity signing (Dilithium), QRNG token, optional QKD \\ 
    \hline

    Initialize Session & Global Coordination Layer \newline Quantum Cryptographic Stack &
    Orchestrator: Identity Enforcer, Key Registrar &
    PQC-based key provisioning, QRNG-derived session ID, quantum entropy seeding \\ 
    \hline

    Secure Client Request & Global Coordination Layer \newline Perception \& Input Interface &
    Orchestrator: Input Validator \newline Agent: Input Filter Module &
    PQC TLS, Dilithium client signature check, QRNG request hash, PII detection and redaction \\ 
    \hline

    Build Task Graph & Global Coordination Layer \newline Planning and Reasoning Engine \newline Securing MCP-Agent Communication &
    Orchestrator: Task Planner, DAG Compiler \newline Agent: Orchestration Router &
    QRNG for DAG node IDs, signed edge policies via Dilithium, Kyber-secured task routing \\ 
    \hline

    Secure Agent Execution & Agent Layer: \newline RAG Module \newline Memory Store \newline Tools Interface \newline MCP Channel &
    Agent: Tool Invoker, Retriever, Memory Engine &
    Agent-to-agent communication via PQC (Kyber), per-agent QRNG token, encrypted tool calls \\ 
    \hline

    Merge And Reason & Memory Store \newline Planning and Reasoning Engine &
    Agent: Reasoning Graph Engine, Belief Updater &
    QRNG DAG merging entropy, Dilithium-signed trace of logic chain, encrypted memory pointers \\ 
    \hline

    Finalize And Respond & Global Coordination Layer &
    Orchestrator: Response Finalizer, Signature Notary &
    Kyber/AES-GCM for final payload encryption, QRNG-based response ID, region-specific sealing \\ 
    \hline

    Record Audit Trail & Global Coordination Layer \newline Quantum Cryptographic Stack &
    Orchestrator: Ledger Writer \newline Agent: Event Emitter &
    Append-only PQC log, QRNG event IDs, QKD-replicated ledger (where available), digital receipts \\ 
    \hline
    \end{tabularx}
    \label{tab:Table3}
\end{table}

Table B.2 presents a representative mapping of some of the key application domains to example use cases where quantum-secured multi-agent systems can be deployed. It highlights the role of agents, cryptographic control mechanisms, and deployment environments in each domain.

\begin{table}[H]
\renewcommand{\thetable}{B.2} % set the display number
\centering
\caption{Quantum-secured multi-agent use cases across domains.}
\label{tab:use_cases}
%\resizebox{\textwidth}{!}{%
%\begin{tabularx}{\textwidth}{|p{.15cm}|X|X|X|X|X|X|X|}
\begin{tabularx}{\textwidth}{p{.15cm}XXXXXXX}
\hline
\textbf{\#} & \textbf{Domain} & \textbf{Use Case}  & \textbf{Agent Roles} & \textbf{PQC Usage} & \textbf{QRNG Usage} & \textbf{QKD Usage} \\
\hline
\hline
1  & Finance \& Banking & Decentralized Financial Infrastructure &  Trading bots, Compliance agents, Arbitrage agents & Digital signatures, Transaction integrity & Pricing tokens, Trade nonces & Central bank $\leftrightarrow$ Exchange security \\
\hline
2  & Aerospace \& Space & Space-Earth Autonomy & Telemetry processors, Navigation agents, Signal relayers & Signed commands, Fault tolerance protocols & Command entropy, Data integrity tokens & Earth $\leftrightarrow$ satellite laser-based QKD \\
\hline
3  & Healthcare \& Life Sciences & Global Health AI & Federated trainers, Diagnosis support agents & Model update encryption, GDPR/HIPAA compliance & Federated round key generation & Optional: Hospital-to-hospital secure sync \\
\hline
4  & Energy \& Utilities & Smart Grids \& Autonomous Utilities & Relay controllers, Load balancers, Demand forecasters & Telemetry protection, Load switch verification & Spoof prevention, Power dispatch randomness & Cross-grid or regional coordination QKD \\
\hline
5  & Cybersecurity \& Intelligence & Intelligence \& Threat Hunting  & Threat modelers, SOC analysts, Packet inspectors            & Threat signature validation, Data sealing      & Forensic log entropy, Attack trace prevention & NATO or inter-agency classified coordination \\
\hline
6  & Transportation \& Logistics & Secure Global Supply Chains & Asset trackers, Route planners, Customs clearance agents & Shipment signing, Smart contract enforcement & Unique tag tokens, Anti-counterfeiting & Port authority $\leftrightarrow$ regulatory coordination \\
\hline
7  & Legal \& Compliance & Autonomous Regulation Agents & Law enforcement bots, Policy engines, Transparency auditors & Regulation signing, Action auditability & Redaction traceability, Decision randomness & Governance node interlinks \\
\hline
8  & Defense \& Homeland Security & Autonomous Defense Coordination & Recon agents, Swarm controllers, Defense planners & Control link protection, Event sealing & Random mission IDs, Time-sensitive keys & Tactical interlinking between allied nodes \\
\hline
9  & Smart Manufacturing & AI-Driven Factory Orchestration & Scheduler bots, Inventory agents, Quality controllers & IoT device authentication, Workflow signing    & Random order tags, Inventory shielding & Multi-factory secure updates \\
\hline
10 & Government \& Public Sector   & Sovereign AI Decision Systems & Policy modelers, Social risk assessors & Secure messaging, Decision validation & Risk scenario simulation, Decision entropy    & Government cloud $\leftrightarrow$ shared agency secure links \\
\hline
11 & Telecommuni-cations & Quantum-Secure Edge Networking & Network optimizers, Bandwidth allocators & Session handoff encryption, E2E signature auth & Nonce generation for edge packets    & Metro-to-core fiber QKD \\
\hline
13 & Climate \& Sustainability & Distributed Environmental Monitoring  & Climate modelers, Sensor readers, Risk forecasters & Data signature, Model authenticity & Random sampling tokens, Sensor integrity & Cross-border agency communications (e.g., UNEP, NOAA) \\
\hline
14 & Retail \& e-Commerce & Autonomous Shopping \& Fulfillment & Inventory pickers, Price trackers, Recommendation engines & Order verification, Pricing trace & Buyer preference entropy, ID tokenization & High-value inter-warehouse coordination\\
\hline

\end{tabularx}%
%}
\end{table}

%Security Dimension Comparison 
%% table
%\addtocounter{table}{-1}
\begin{table}[h]
\renewcommand{\thetable}{B.3} % set the display number
\centering
\caption{Comparison of Security Dimensions for Classical, PQC, and QKD+QRNG Systems}
%\renewcommand{\arraystretch}{1.4}
%\begin{tabularx}{\textwidth}{|X|X|X|X|}
\begin{tabularx}{\textwidth}{XXXX}
\hline
\textbf{Security Dimension} & \textbf{Classical Crypto} & \textbf{Post-Quantum Crypto (PQC)} & \textbf{QKD + QRNG (Quantum-Resilient)} \\
\hline
\hline
Resistance to Quantum Attacks & Low – Vulnerable & High – based on lattice, hash, or code hardness \cite{ref_sec6.5_table(1-3_4-3)_Scrivano2025} \cite{ref_sec6.5_table(1-3)_cisco_Gill2025Cisco} & Highest – Information-theoretic security; passive detection of eavesdropping \cite{ref_6.5_table(1-4)_NSA_QKD}\\
\hline
Forward Secrecy Longevity & Moderate & Strong (at PQC level) & Superior – symmetric keys refreshed via QKD \\
\hline
Entropy Assurance & Weak – PRNG risk & Better, but software-limited & Physical-grade entropy via QRNG, near-ideal unpredictability\\
\hline
Key Distribution Infrastructure & Universal (IP/TLS) & Fully compatible (TLS-PQC-ready) \cite{ref_6.5_table(4-3_7-3)_Demir2025PQCDeployment} \cite{ref_sec6.5_table(1-3_4-3)_Scrivano2025} & Requires dedicated hardware (fiber/satellite), limited reach\\
\hline
Vulnerability to Eavesdrop/Replay & High & Low & Very Low – quantum disturbances detectable \\
\hline
Implementation Complexity & Low & Moderate & Very High – complex hardware and calibration requirements \\
\hline
Scalability \& Latency & High & High (key exchange $\approx$ 1–5 ms) \cite{ref_6.5_table(4-3_7-3)_Demir2025PQCDeployment} \cite{ref_6.5_table(7-3)_Dong2025EPQUIC} & Low – setup delays in hundreds of ms/sec; limited throughput\\
\hline
Standardization \& Compliance & Mature & Emerging, NIST-approved (Kyber, Dilithium) \cite{ref_6.5_table(8-3)_NIST2024PQCStandards} & Niche adoption; few operational deployments\\
\hline
\end{tabularx}
\label{tab:Table6}
\end{table}

%%% Appendix C
{\textbf{Appendix C — Empirical validation details} \\}

This subsection provides the detailed empirical validation material that complements the compressed main-text evaluation section. In alignment with the two-phase validation strategy summarized in the main paper, the supplementary material presents the full experimental architecture, deployment protocol, hardware and cryptographic setup, local cryptographic micro-benchmarks, cloud-scale integration analysis, and adversarial resilience results. Together, these details further substantiate the operational feasibility, secure execution across distributed trust boundaries, and scalable behavior of the proposed quantum-secure-by-construction framework under both controlled and production-oriented conditions.

To ensure rigorous evaluation, the framework was validated through a dual-phase experimental strategy. Phase I utilized a controlled local testbed to isolate the computational latency of cryptographic primitives, while Phase II expanded this into a production-grade cloud deployment on Azure to validate system scalability and network latency integration.\\

\noindent\rule{\textwidth}{0.4pt}
%\hline 
\textbf{\\ C1: Experimental testbed architecture and protocol}\\
\noindent\rule{\textwidth}{0.4pt}
%\hline
%%\subsubsection{Experimental testbed architecture and protocol}
%%\label{sec:architecture}

The system architecture enforces a Zero-Trust model where all components, regardless of their hosting environment, communicate exclusively via quantum-secured channels (Figure~\ref{fig:logical_architecture}). The deployment topologies are defined as follows.

\begin{figure}[!htbp]
\setcounter{figure}{0}
\renewcommand{\thefigure}{C.1}
    \centering
    \includegraphics[width=0.6\textwidth]{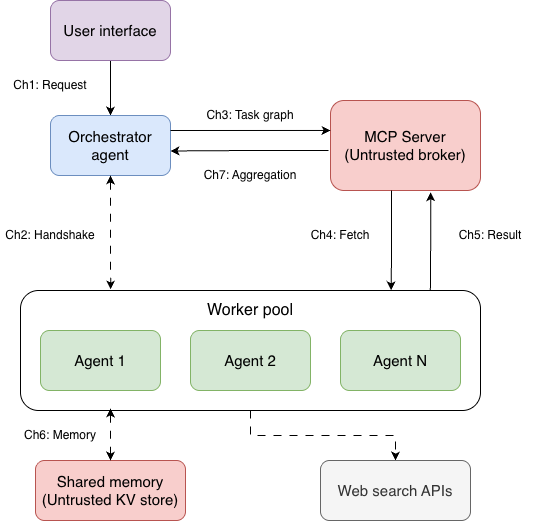}
    \caption{Logical System Architecture: A Zero-Trust, quantum-secured multi-agent system enforcing cryptographic boundaries across seven critical communication channels.}
    \label{fig:logical_architecture}
\end{figure}

\begin{itemize}
    \item \textbf{Local Research Testbed:} Designed for micro-benchmarking, this environment isolates the CPU cycles required for PQC key encapsulation and signing, utilizing a direct hardware interface to a Quantum Random Number Generator (QRNG) for entropy injection.
    \item \textbf{Cloud-Native Sidecar Deployment (Azure):} To validate real-world applicability, the architecture was deployed to the Azure East US region using a Sidecar Pattern (Figure~\ref{fig:azure_architecture}). Each architectural component, including the Orchestrator, Agents, MCP, and Shared Memory, is deployed as a container group paired with a custom Nginx-OQS ingress/egress gateway. This sidecar handles all PQC-TLS 1.3 termination and establishment, creating a transparent mesh of quantum-secured tunnels without requiring modification to the core agent logic.
\end{itemize}

%\begin{figure}[!htbp]
\begin{figure}[H]
\setcounter{figure}{0}
\renewcommand{\thefigure}{C.2}
    \centering
    \includegraphics[width=0.5\textwidth]{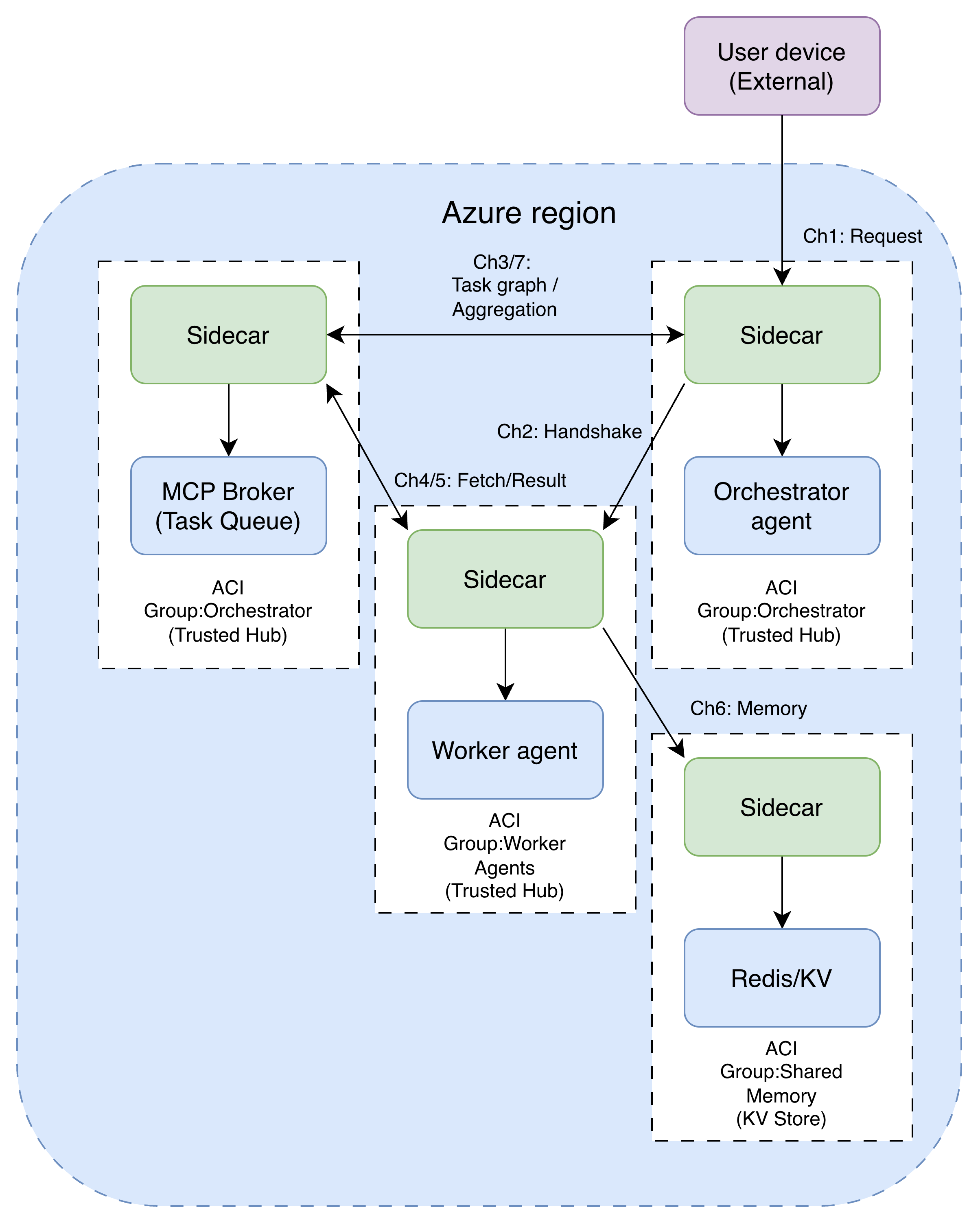}
    \caption{Azure PQC Sidecar Architecture: Securing inter-agent communication via transparent PQC-TLS tunnels.}
    \label{fig:azure_architecture}
\end{figure}

The operational workflow is defined by the trust boundary loop, comprising seven critical channels that necessitate quantum-resistant protection (Table~\ref{tab:7_channels_final_compressed}). The MCP Server and Shared Memory are explicitly modeled as untrusted brokers residing on the public or shared network.

The protocol execution follows four key phases:
\begin{enumerate}
    \item \textbf{Initialization (Channels 1--2):} The user submits a query authenticated with a QRNG nonce and PQC signature. Worker agents register with the Orchestrator via PQC-signed handshakes.
    \item \textbf{Orchestration (Channel 3):} The Orchestrator decomposes the query into a signed task graph and publishes it to the untrusted MCP Server, encrypted via PQC-KEM or QKD.
    \item \textbf{Execution (Channels 4--6):} Agents fetch tasks, verify the Orchestrator's signature, and execute operations. Inter-agent collaboration via Shared Memory is secured using ephemeral PQC session keys.
    \item \textbf{Aggregation (Channel 7):} The Orchestrator retrieves results, sequentially verifying the PQC signature and decrypting each agent's payload to compile the final report.
\end{enumerate}

For data confidentiality across all channels, the system prioritizes QKD when the simulated link is active; otherwise, it automatically falls back to PQC-based key encapsulation mechanisms to ensure persistent quantum safety.\\

%\hline
\noindent\rule{\textwidth}{0.4pt}
\textbf{\\ C2: Experimental setup and cryptographic parameters}\\
\noindent\rule{\textwidth}{0.4pt}
%\hline

%\subsubsection{Experimental setup and cryptographic parameters}

The experiments were conducted using a dedicated external QRNG device. The computing environment specifications are detailed in Table~\ref{tab:comp_env_specs}. The quantum entropy source used was the Quantis USB 4M QRNG \cite{IDQ2025}, the specifications of which are listed in Table~\ref{tab:qrng_specs}.

%\begin{table}[h!]
%\renewcommand{\thetable}{B.1} % set the display number
%\centering
%\caption{Quantum-secured multi-agent use cases across domains.}
%\label{tab:use_cases}
%\resizebox{\textwidth}{!}{%
%\begin{tabularx}{\textwidth}{|p{.15cm}|X|X|X|X|X|X|X|}
%\hline

\begin{table}[h]
\renewcommand{\thetable}{C.1} % set the display number
\centering
\caption{Computing Environment Specifications}
\label{tab:comp_env_specs}
\scriptsize
\setlength{\tabcolsep}{4pt}
\begin{tabular}{ll}
\toprule
\textbf{System Specification} & \textbf{Value} \\
\midrule
\hline
Processor & Apple M3 Pro, 3.8 GHz, 12 cores \\
Memory (RAM) & 36 GB \\
Operating System & macOS 15.2 \\
\bottomrule
\end{tabular}
\end{table}

\begin{table}[h]
\renewcommand{\thetable}{C.2} % set the display number
\centering
\caption{ID Quantique Quantis USB 4M QRNG Specifications}
\label{tab:qrng_specs}
\scriptsize
\setlength{\tabcolsep}{4pt}
\begin{tabular}{ll}
\toprule
\textbf{Specification} & \textbf{Value} \\
\midrule
\hline
Random Bit Rate & 4 Mbit/s $\pm$ 10\% \\
Thermal Noise Contribution & $< 1\%$ \\
Storage Temperature & -25$^\circ$C to +85$^\circ$C \\
USB Specification & 2.0 \\
Power & Via USB port \\
\bottomrule
\end{tabular}
\end{table}

This device delivers raw quantum bits predominantly based on quantum vacuum fluctuations, ensuring a high-quality entropy source.\\

%\hline
\noindent\rule{\textwidth}{0.4pt}
\textbf{\\ C3: Methodology and experimental rationale}\\
\noindent\rule{\textwidth}{0.4pt}
%\hline
%\subsubsection{Methodology and experimental rationale}
%\label{sec:methodology}
The central tenet of this research is that the computational overhead of post-quantum cryptography (PQC) and quantum entropy injection is a manageable trade-off for ensuring long-term security in distributed agent systems. To validate this, our methodology is divided into two phases.

\begin{itemize}
    \item \textbf{Phase I: Cryptographic Micro-Benchmarking (Isolation):} This phase quantifies the raw computational cost of quantum-safe primitives compared to a classical baseline. By running these tests on a local testbed with direct QRNG hardware access, we isolate algorithm execution time from network variance.
    \item \textbf{Phase II: Cloud Integration Analysis (Contextualization):} This phase measures the total transaction latency of the full system when deployed in a production cloud environment. This step is critical to demonstrate that, in real-world distributed systems, the security overhead of PQC is largely masked by the substantially higher latencies of cloud networking and AI inference.
\end{itemize}

The experiment specifically selected seven communication channels that collectively form the full interactive trust boundary loop in Figure~\ref{fig:Figure1}. The MCP Server and Shared Memory components are modeled as an untrusted central broker residing on the global network, making cryptographic protection on these links essential. We intentionally omitted purely internal non-networked hand-off links whose cryptographic cost is already captured by the sequenced main channels.\\\\

%\hline
\noindent\rule{\textwidth}{0.4pt}
\textbf{\\ C4: Security resilience} \\
\noindent\rule{\textwidth}{0.4pt}
%\hline
%\subsubsection{Security resilience}

To statistically validate the defense mechanisms, we conducted a large-scale Monte Carlo adversarial simulation with $100{,}000$ injected malicious vectors. The dataset was divided between payload tampering attempts and replay attacks using stale nonces.

As detailed in Table~\ref{tab:security_resilience_final}, the system achieved a 100\% rejection rate. The PQC layer invalidated all tampered payloads during signature verification, while the QRNG layer detected all replay attempts by enforcing unique high-entropy nonces for each transaction. These results confirm the separation of security duties across the architecture, with PQC providing integrity and QRNG providing freshness.

\begin{table}[H]
\renewcommand{\thetable}{C.3} % set the display number
\centering
\caption{Adversarial Simulation Results ($N=100{,}000$ Vectors)}
\label{tab:security_resilience_final}
\small
\setlength{\tabcolsep}{6pt}
\begin{tabular}{l c c c}
\toprule
\textbf{Attack Vector} & \textbf{Injected} & \textbf{Detected/Blocked} & \textbf{Efficacy} \\
\midrule
\hline
Integrity Violation (Tamper) & 50,000 & 50,000 & 100.0\% \\
Replay/Freshness Attack & 50,000 & 50,000 & 100.0\% \\
\midrule
\textbf{Total} & \textbf{100,000} & \textbf{100,000} & \textbf{100.0\%} \\
\bottomrule
\end{tabular}
\end{table}

\end{document}